\documentclass[twoside,11pt]{article}
\usepackage[dvipsnames]{xcolor}
\usepackage{style_file}
\usepackage{amsmath}
\usepackage{algorithm}
\usepackage[noend]{algpseudocode}
\usepackage{graphicx}
\usepackage{subcaption}
\usepackage{xurl}

\newcommand{\R}{{\mathbb R}}

\ShortHeadings{BNP for Neural Network Training}{Lange, Helfrich, and Ye}
\firstpageno{1}

\begin{document}\thispagestyle{plain}

\title{Batch Normalization Preconditioning for Neural Network Training}

\author{\name Susanna\ Lange \email susanna.lange@uky.edu      \\
\addr Department of Mathematics,  \\
       University of Kentucky \\
       Lexington, KY 40506
\AND
\name Kyle \ Helfrich \email kehelfrich@owu.edu \\
       \addr Department of Mathematics and Computer Science,  \\
       Ohio Wesleyan University \\
       Delaware, OH 43015
       \AND
       \name Qiang\ Ye  \email qiang.ye@uky.edu \\
       \addr Department of Mathematics,  \\
       University of Kentucky \\
       Lexington, KY 40506}

\maketitle

\begin{abstract}
Batch normalization (BN) is a popular and ubiquitous method in deep learning that has been shown to decrease training time and improve generalization performance of neural networks. Despite its success, BN is not theoretically well understood. It is not suitable for use with very small mini-batch sizes or online learning. In this paper, we propose a new method called Batch Normalization Preconditioning (BNP). Instead of applying normalization explicitly through a batch normalization layer as is done in BN, BNP applies normalization by conditioning the parameter gradients directly during training. This is designed to improve the Hessian matrix of the loss function and hence convergence during training. One benefit is that BNP is not constrained on the mini-batch size and works in the online learning setting.
Furthermore, its connection to BN provides theoretical insights on how BN improves training and how BN is applied to special architectures such as convolutional neural networks. For a theoretical foundation, we also present a novel Hessian condition number based convergence theory for a locally convex but not strong-convex loss, which is applicable to networks with a scale-invariant property.
\end{abstract}

\begin{keywords}
  Deep neural networks, Convolutional neural networks, Preconditioning, Batch Normalization
\end{keywords}

\section{Introduction}

Batch normalization (BN) is one of the most widely used techniques to improve neural network training. It was originally designed in \cite{Ioffe15} to address \textit{internal covariate shift}.  It has been found to  
increase network robustness with respect to parameter and learning rate initialization, to decrease   training times, and to improve network regularization. 
Over the years, BN has become a standard technique  in deep learning but the theoretical understanding of BN is still limited.   
There have been many papers that analyze various properties of BN but some important issues remain unaddressed. 
 


In this paper, we develop a method called {\em Batch Normalization Preconditioning} (BNP).   Instead of using mini-batch statistics to  normalize hidden variables as in BN, BNP uses these statistics to  transform the trainable parameters through \textit{preconditioning},  which improves the conditioning of the Hessian of the loss function and hence accelerates convergence.  This is implicitly done by applying a transformation on the gradients during training. Preconditioning is a general technique in numerical analysis \citep{10.5555/829576} that uses a parameter transformation  to accelerate convergence of iterative methods. Here, we show  that potentially large differences in variances of  hidden variables over a mini-batch adversely affect the conditioning of the Hessian. We then derive a suitable parameter transformation that is equivalent to normalizing the hidden variables. 

Both BN and BNP use mini-batch statistics to accelerate convergence but the key difference is that BNP 
does not change the network architecture,
but instead uses the mini-batch statistics in an equivalent transformation of parameters to accelerate convergence. In particular, BNP may use statistics computed from a larger dataset rather than the mini-batches. 
In contrast, BN has the mini-batch statistics embedded in the architecture. This has the advantage of allowing gradients to pass through these statistics during training but this is also a theoretical disadvantage because the training network is dependent on the mini-batch inputs. In particular, the inference network (with a single input) is different from the training network (with mini-batch inputs).  
This  may pose a challenge in implementation with small mini-batches as well as in analysis, even though BN largely works well in practice. 
We outline 
our main contributions below: 

\begin{itemize}

\item We develop a novel Hessian condition number based convergence theory for a locally convex but not strong-convex loss, which is applicable to  networks with a scale-invariant property (\cite{NIPS2015_eaa32c96,meng2018gsgd}).

\item Under certain conditions, BNP is  equivalent to BN. 
Because of this equivalency, the theoretical basis of BNP provides an explanation of why BN works. 
  
\item While BNP 
has a performance comparable to BN in general, it  outperforms BN in the situation of very small mini-batch sizes or online learning where BN is known to have difficulties. 

\item  BN has been adapted to special architectures like convolutional neural networks (CNNs)  \citep{Ioffe15} and recurrent neural networks (RNNs) \citep{Laurent_2016,cooijmans2016recurrent} but the computation of the required mini-batch statistics is more nuanced.  For example, in CNNs the mean and variance across the mini-batch and spatial dimensions are used without clear justification.  Derivation of BNP for CNNs naturally leads to using these statistics and provides a theoretical understanding of how BN should be applied to CNNs.  
\end{itemize}

Throughout the paper, $\| \cdot \|$  denotes the 2-norm. 
If $A$ is a square and invertible matrix,  $\kappa(A)=\|A\|\|A^{-1}\|$ denotes the condition number of $A$. For a singular or rectangular matrix $A$, we define the condition number as $\kappa(A) =\frac{\sigma_{\text{max}}}{\sigma^*_{\text{min}}}$ , where $\sigma^*_{\text{min}}$ is the smallest nonzero singular value of $A$ and $\sigma_{\text{max}}$ the largest singular value of $A$. 
All vectors refer to column vectors. A diagonal matrix with diagonal entries given by the vector $v$ is denoted by diag$(v)$, and $e=[1, 1, \cdots, 1]^T$ denotes the vector of ones with a suitable dimension.  All functions and  operations of vectors are applied elementwise. 
    
In Section \ref{BN_section}, we present BN and various related works. We
introduce preconditioned gradient descent in Section \ref{pgd_section} followed by a Hessian based convergence theory for convex but not strongly-convex losses in Section \ref{hessian_convergence_section}. We derive the BNP method in Section \ref{BNP_section}. We discuss the connection between BN and BNP in Section \ref{relation_section} and derive BNP for CNNs in Section \ref{CNN_section}. Finally, we provide experimental results in Section  \ref{exp_main}.  Proofs of most results are given in the appendix.

\section{Batch Normalization and Related Work}
\label{BN_section}


Consider a fully connected multi-layer neural network in which the  $\ell$-th hidden layer is defined by  
\begin{equation}
\label{vanilla_eqn}
 h^{(\ell)}= g(W^{(\ell)} h^{(\ell-1)} +b^{(\ell)}), 
\end{equation}
 where $h^{(0)}$ is the network input $x$, $g$ is an elementwise nonlinear function, and $h^{(\ell)} \in \mathbb{R}^{n_{\ell}}$ is the $\ell$-th hidden variable with $W^{(\ell)}$ and $b^{(\ell)}$ being the associated weight and bias.  We refer to such a network as a \textit{vanilla network}.   
During training, let $\{x_1, x_2, \hdots, x_N\}$ be a mini-batch consisting of $N$ examples and $H = [h^{(\ell-1)}_1, h^{(\ell-1)}_2, \hdots, h^{(\ell-1)}_N]^T$ the matrix of associated hidden variables of layer $\ell-1$.  
BN replaces the  $\ell$-th hidden layer by 
\begin{equation}
\label{post_eqn}
h^{(\ell)}= g\left(W^{(\ell)} \mathcal{B}_{\beta,\gamma}(h^{(\ell-1)}) +b^{(\ell)}\right); \hspace{0.2cm}
\mathcal{B}_{\beta,\gamma}\left( h^{(\ell-1)} \right) = \gamma\frac{h^{(\ell-1)} - \mu_H}{\sigma_H} + \beta
\end{equation}
where $\sigma_H$ and $\mu_H$ are the standard deviation and mean vectors of $\{ h^{(\ell-1)}_i \}$ (i.e. the rows of $H$) and $\gamma,\beta$ are the respective trainable re-scaling and re-centering  parameter vectors. The BN transformation operator $\mathcal{B}_{\beta,\gamma}\left(\cdot \right)$ first normalizes the activation $h^{(\ell-1)}$ and then re-scales and re-centers it with $\gamma$ and $\beta$; see Algorithm 1 for details. In practice, the denominator in (\ref{post_eqn}) is the square root of the variance vector plus a small $\epsilon > 0$ to avoid division by zero. 
Note that BN can be implemented as $h^{(\ell)}= g\left(\mathcal{B}_{\beta,\gamma}(W^{(\ell)} h^{(\ell-1)})\right)$,
which will be referred to as \textit{pre-activation} BN.  In this paper we will focus on (\ref{post_eqn}), which is called  \textit{post-activation} BN. 

 \begin{algorithm}[ht] \label{alg:BN_alg}
 \caption{Batch Normalization $\mathcal{B}_{\beta, \gamma} (h)$}
 \begin{algorithmic}
 \State \textbf{Input:} $h_i\in \mathbb{R}^{n_{\ell}}$ and $H = [h_1, h_2, \hdots, h_N]^T \in \mathbb{R}^{N \times n_{\ell}}$. 
 \State 1. $\mu_{H} \gets \frac{1}{N}\sum_{i=1}^{N}h_i \in \mathbb{R}^{n_{\ell}}$ (Mini-batch mean).
 \State 2. $\sigma^2_{H} \gets \frac{1}{N}\sum_{i=1}^{N}(h_i - \mu_H)^2 \in \mathbb{R}^{n_{\ell}}$ (Mini-batch variance. The square is elementwise).
 \State 3. $h \gets \text{diag}\left(\frac{1}{\sqrt{\sigma^2_{H}+\epsilon}}\right)\left(h - \mu_H \right)$ (Centering and scaling).
 \State \textbf{Output:} $\mathcal{B}_{\beta, \gamma} (h, A) \gets \text{diag}(\gamma) h + \beta $ (Re-scaling and re-shifting).
 \end{algorithmic}%
 \end{algorithm}

During training, the BN network (\ref{post_eqn}) changes with different mini-batches since $\sigma_H$ and $\mu_H$ are based on the mini-batch input.  During inference, one input is fed into the network and $\sigma_H$ and $\mu_H$ are not well-defined.  Instead, the average of the mini-batch statistics $\sigma_H$ and $\mu_H$ computed during training, denoted as $\sigma$ and $\mu$, are used \citep{Ioffe15}. Alternatively, $\sigma$ and $\mu$ can be computed using  moving averages computed as 
$\sigma \leftarrow \rho \sigma + (1-\rho) \sigma_H$ and $\mu\leftarrow \rho \mu + (1-\rho) \mu_H$ for some $0< \rho<1.$

To reconcile the training and inference networks, 
{\em batch renormalization} \citep{ioffe2017batch} applies an additional affine transform in   $\mathcal{B}_{\beta,\gamma}\left(\cdot \right)$ as 
\begin{equation}
\label{renorm}
\mathcal{R}_{\beta,\gamma}\left( h^{(\ell-1)} \right) = \gamma\frac{h^{(\ell-1)} - \mu}{\sigma} + \beta
=\gamma\left( \frac{h^{(\ell-1)} - \mu_H}{\sigma_H} s +d \right) +\beta
\end{equation}
where $s=\sigma_H/\sigma$ and $d=(\mu_H-\mu)/\sigma$. The first formula in (\ref{renorm}) is used during inference, but the second one is used during training where $s,d$ are considered fixed but $\sigma_H$ and $\mu_H$ are considered parameters with gradients passing through them. 

There have been numerous works that analyze different aspects of BN.  The work by \cite{arora2018theoretical} analyzes scale-invariant properties to demonstrate BN's robustness with respect to learning rates. \cite{DBLP:journals/corr/abs-1709-09603} explores the same property in a Riemannian manifold optimization. The empirical study performed by \cite{Bjorck18} indicates how BN allows for larger learning rates which can increase implicit regularization through random matrix theory.  In \cite{Cai19} and \cite{Kohler18}, the ordinary least squares problem is examined and  convergence properties are proved.
\cite{Kohler18} also discusses  
a two-layer model.  \cite{Ma19} analyzes some special two-layer models and introduces a method that updates the parameters in BN by a diminishing moving average.   \cite{Santurkar18} proves bounds to show that BN improves Lipschitzness of the loss function and boundedness of the Hessian.  \cite{Lian18} analyzes BN performance on smaller mini-batch sizes and also suggests that batch normalization performs well by improving the condition number of the Hessian. \cite{Yang19} uses a mean field theory to analyze gradient growth as the depth of a BN network increases, and \cite{galloway2019batch}  shows empirically that a BN network is less robust to small adversarial perturbations. \cite{DBLP:journals/corr/Laarhoven17b} discusses the relation between BN and $L_2$ regularization and \cite{DBLP:journals/corr/abs-1809-00846} gives a probabilistic analysis of BN's regularization effects. Additionally, \cite{daneshmand2020batch} provides theoretical results to show that a very deep linear vanilla networks leads to rank collapse, where the rank of activation matrices drops to 1, and BN avoids the rank collapse and benefits the training of the network. Compared with the existing works, our results apply to  general neural networks and our theory has practical implications on how to apply BN to CNNs or on  small mini-batch sizes. 

There are several related works preceding BN. Natural gradient descent \citep{10.1162/089976698300017746} is gradient descent  
in a Riemannian space that can be regarded as preconditioning for the Fisher matrix. More practical algorithms with various approximations of the Fisher matrix have been studied \citep[see][and references contained therein]{Raiko12,Grosse15,Martens16}.
In  \cite{Desjardins15}, a BN-like transformation is carried out to whiten each hidden variable 
to improve the conditioning of the Fisher matrix. Using an approximate Hessian/preconditioner has been explored in \cite{becker:improving}.
\cite{li2018preconditioner} discusses preconditioning for both the Fisher and the Hessian matrix and suggests several possible preconditioners for them. 
\cite{DBLP:journals/corr/abs-1802-06502} and \cite{zhang2017blockdiagonal} use some block approximations of the Hessian, and
\cite{osher2018laplacian} uses a discrete Laplace matrix. Although BNP uses the same framework of preconditioning, the preconditioner is explicitly constructed using batch statistics and we demonstrate how it improves the conditioning of the Hessian.  

There are other normalization methods such as LayerNorm \citep{ba2016layer,Xu19}, GroupNorm \citep{wu2018group}, instance normalization \citep{ulyanov2016instance}, weight normalization \citep{Salimans16}, and SGD path regularization \citep{NIPS2015_5797}. They are concerned with normalizing a group of hidden activations, which are totally independent of our approach as well as BN's. Indeed, they can be combined with BN as shown in \cite{Summers2020Four} as well as our BNP method. 


\section{Batch Normalization Preconditioning (BNP)}
In this section we develop our main preconditioning method for fully connected networks and present its relation to BN.
\subsection{Preconditioned Gradient Descent}
\label{pgd_section}
 Given the parameter vector $\theta$ of a neural network and a loss function $\mathcal{L}=\mathcal{L}(\theta)$, the gradient descent method updates an approximate minimizer $\theta_k$ as:
\begin{equation}\label{eq:gd}
     \theta_{k+1} \leftarrow \theta_k - \alpha \nabla \mathcal{L} (\theta_k), 
\end{equation}
 where $\alpha$ is the learning rate. Here and throughout, we assume $\mathcal{L}$ is twice continuously differentiable, i.e. $\nabla^2 \mathcal{L}(\theta)$ exists and is continuous. Let $\theta^{*}$ be a local minimizer of $\mathcal{L}$ (or $\nabla \mathcal{L}(\theta^*)=0$ and the Hessian matrix $\nabla^2 \mathcal{L}(\theta^*)$ is symmetric positive semi-definite) and let $\lambda_{\text{min}}$ and $\lambda_{\text{max}}$ be the minimum and maximum eigenvalues of  $\nabla^2 \mathcal{L} (\theta^*)$. Assume $\lambda_{\text{min}} > 0$. Then for any $\epsilon>0$, there is a  small neighborhood around $\theta^{*}$ such that, for any initial approximation $\theta_0$ in that neighborhood, we have 
 \begin{equation}\label{eq:oneiter}
 \| \theta_{k+1} - \theta^{*} \| \leq (r+\epsilon)  \|\theta_{k} - \theta^{*} \|,
\end{equation}
where $r = \max \{ \left|1 - \alpha \lambda_{\text{min}} \right|, \left|1 - \alpha \lambda_{\text{max}} \right| \}$; see \cite{Polyak1964SomeMO} and \cite[Example 4.1]{10.5555/829576}. In order for $r<1$, we need $\alpha < 2/\lambda_{\text{max}} = 
{2}/{\|\nabla^2_{\theta}\mathcal{L}(\theta^{*}) \|}$.  Moreover, minimizing $r$ with respect to $\alpha$ yields $\alpha = \frac{2}{\lambda_{\text{min}} + \lambda_{\text{max}}}$ with optimal convergence rate:
\begin{equation}
  r = \frac{\kappa - 1}{\kappa + 1},    \label{r_equation}\end{equation} 
where
  \begin{equation*} \kappa= \kappa \left(\nabla^2 \mathcal{L} (\theta^*)\right) =\frac{\lambda_{\text{max}}}{\lambda_{\text{min}}}
    \label{kappa_eqn}
\end{equation*}
is the condition number of $\nabla^2 \mathcal{L} (\theta^*)$ in 2-norm.
Thus the optimal rate of convergence is determined by the condition number of the Hessian matrix.  

To improve convergence, we consider a change of variable $\theta = Pz$ for the loss function $\mathcal{L} = \mathcal{L}(\theta)$, which we call a {\em preconditioning} transformation. Writing $\mathcal{L} = \mathcal{L}(Pz)$, the gradient descent in $z$ is
\begin{align}
    z_{k+1} = z_{k} - \alpha \nabla_{z}\mathcal{L}\left(Pz_k \right) = z_k - \alpha P^{T}\nabla_{\theta}\mathcal{L}\left(\theta_k \right),
    \label{pgd_sgd}
\end{align}
where $\theta_k=P z_k$.
 Let $z^{*}=P^{-1}\theta^*$. The corresponding convergence bound becomes
$$\left\lVert z_{k+1} - z^{*} \right\rVert \leq (r+\epsilon) \left\lVert z_{k} - z^{*} \right\rVert $$ with
$r$ as determined by (\ref{r_equation}) and the Hessian of $\mathcal{L}$ with respect to $z$:
$$\nabla^{2}_{z}\mathcal{L}\left(Pz^{*}\right) = P^{T}\nabla^{2}_{\theta}\mathcal{L}\left({\theta}^{*}\right)P.$$  If $P$ is such that $P^{T}\nabla^{2}_{\theta}\mathcal{L}\left({\theta}^{*}\right)P$ has a better condition number than $\nabla^{2}_{\theta}\mathcal{L}\left({\theta}^{*}\right)$,  $r$ is reduced and the convergence accelerated.
Multiplying (\ref{pgd_sgd}) by $P$,  we obtain the equivalent update scheme in $\theta_k=P z_k$:
\begin{align}
    \theta_{k+1} = \theta_{k} - \alpha PP^{T}\nabla_{\theta}\mathcal{L}\left(\theta_{k}\right).
    \label{pd_eqn3}
\end{align}
We call $PP^{T}$  a preconditioner. Note that (\ref{pd_eqn3}) is an implicit implementation of the iteration  (\ref{pgd_sgd}) and simply involves modifying the gradient in the original gradient descent (\ref{eq:gd}). 
\subsection{Hessian Based Convergence Theory for Neural Networks}
\label{hessian_convergence_section}
The Hessian based convergence theory given above assumes strong local convexity at a local minimum (i.e. $\lambda_{\min} >0$). This may not hold for a neural network (\ref{vanilla_eqn}). For example, if we use the ReLU nonlinearity in  (\ref{vanilla_eqn}), the network output is invariant under simultaneous scaling of the $j$th row of $W^{(\ell-1)}$ by any $\alpha > 0$ and the $j$th column of $W^{(\ell)}$ by $1/\alpha$. This \emph{positively scale-invariant property} has been discussed in \cite{NIPS2015_eaa32c96} and \cite{meng2018gsgd} for example and is known to lead to a singular Hessian.
Here, we develop a generalization of the strong convexity based theory (\ref{eq:oneiter}) to this situation.

For a fixed network parameter $\theta_0$, we can write all its invariant scalings for all possible rows and columns in the network as $ \theta_0 (t)$ where $0 < t \in \R^k$ for some $k$ is a vector of scaling and $\theta_0 (t_0) =\theta_0$ for some $t_0$. We call  $ \theta=\theta_0 (t)$ a {\em positively scale-invariant manifold} at $ \theta_0$. Then, the network output is constant for $\theta$ on a positively scale-invariant manifold in the parameter space. 
This implies that the loss function $\mathcal{L} (\theta_0 (t))=\frac{1}{N}\sum_{i=1}^N L\left(f(x_i, \theta_0 (t)), y_i\right)$ is constant with respect to $t$, where $f(x_i, \theta_0 (t))$ is the output of the network with the parameter $\theta_0 (t)$ and input $x_i$.

If $\theta^*$ is a local minimizer, let  $ \theta^* (t)$ be the positively scale-invariant manifold at $ \theta^*$. Then $\theta^* (t)$ is a local minimizer for all $t>0$. So $\nabla \mathcal{L} (\theta^* (t))=0$ for all $t>0$ and thus, by taking derivative in $t$, $\nabla^2 \mathcal{L} (\theta^* (t)) D_t \theta^*(t)=0$, where $D_t \theta^*(t) =[\frac{\partial \theta^*_i (t)}{\partial t_j}]_{i,j}$ denotes the Jacobian matrix of $\theta^* (t)$ with respect to $t$.   Thus, the Hessian $\nabla^2 \mathcal{L} (\theta^* (t))$ is singular with the null space containing at least the column space ${\cal C}(t):= {\rm Col}\left( D_t \theta^*(t)\right)$.

With $\lambda_{\min}=0$, the condition number based convergence theory (\ref{eq:oneiter}) does not apply.
However, the theory can be generalized by considering the error $\theta_k - \theta^*_k$, where $\theta^*_k$ is the local minimizer closest to $\theta_k$, i.e. $\theta^*_k=\theta^* (t_k)$ where $t_k=\text{argmin}_{t>0} \|\theta_k - \theta^* (t)\|.$ By taking derivative, we have $(\theta_k - \theta^*_k)^T D_t \theta^* (t_k)=0$, i.e. $\theta_k - \theta^*_k$ is orthogonal to the tangent space, or $\theta_k - \theta^*_k \in {\cal C}(t_k)^\perp$. With this, we can prove (see Appendix) 
\[
\|\theta_{k+1} - \theta^*_{k+1}\| \le \|\theta_{k+1} - \theta^*_k\|
\le \|(I-\alpha \nabla^2 \mathcal{L} (\theta^*_k)) (\theta_k - \theta^*_k)\| +O \left(\|(\theta_k - \theta^*_k)\|^2\right).
\]
and thus a bound similar to (\ref{eq:oneiter}) with $\lambda_{\min}$ replaced by $\lambda^*_{\min}$, the smallest eigenvalue of $\nabla^2 \mathcal{L} (\theta^*_k)$ on ${\cal C}(t_k) ^\perp$. Assuming  ${\cal C}(t_k)$ is exactly the null space of $\nabla^2 \mathcal{L} (\theta^*_k)$, i.e. the singularity of the Hessian is entirely due to the positive scale-invariant property,  we have $\lambda^*_{\min}>0$ being the smallest nonzero eigenvalue. We can then analyze convergence in terms of $\lambda^*_{\min}.$
We present a complete result in the following theorem with a proof given in the Appendix.
\begin{theorem}
\label{thm:hessian_convergence}
Consider a loss function $\mathcal{L}$ with continuous third order derivatives and with a positively scale-invariant property. If $\theta=\theta^*(t)$ is a positively scale-invariant manifold at local minimizer $\theta^*$, then the null space of the Hessian $\nabla^2 \mathcal{L} (\theta^* (t))$   contains at least the column space ${\rm Col}\left( D_t \theta^*(t)\right)$. Furthermore, for the gradient descent iteration $\theta_{k+1}=\theta_{k}-\alpha \nabla \mathcal{L} (\theta_k)$, let $\theta^*_k$ be the local minimizer closest to $\theta_k$, i.e. $\theta^*_k=\theta^* (t_k)$ where $t_k=\text{argmin}_{t>0} \|\theta_k - \theta^* (t)\|$ and assume that  the null space of $\nabla^2 \mathcal{L} (\theta^* (t))$ is equal to ${\rm Col}\left( D_t \theta^*(t)\right)$. Then, for any $\epsilon >0$, if our initial approximation $\theta_0$ is such that $||\theta_0 - \theta^*_0||$ is sufficiently small, then  we have, for all $k \geq 0$,
$$||\theta_{k+1}-\theta^*_{k+1} ||\leq (r+\epsilon)||\theta_k-\theta^*_k||,$$ where
$r=\max\{|1-\alpha \lambda^{*}_{min}|, ||1-\alpha \lambda_{max}||\}$ 
and $\lambda^*_{min}$ and $\lambda_{max}$ are the smallest nonzero eigenvalue and the largest eigenvalue of $\nabla^2 \mathcal{L} (\theta^*_k)$, respectively.
\end{theorem}

As before, the optimal $r$ is given by (\ref{r_equation}) with a condition number for the singular Hessian defined by $\kappa^* = \frac{\lambda_{max}}{\lambda^*_{min}}$. This eliminates the theoretical difficulties of the Hessian singularity caused by the scale-invariant property. 

We note that there are convergence results in less restrictive settings, for instance on L-Lipschitz  or $\beta$-smoothness functions; see \cite{robust_stoch_approx} and \cite{bubeck2015convex}. They are applicable to convex but not strong-convex functions but they only imply convergence of order $O(1/k)$. Without a linear convergence rate, it is not clear whether they can be used for convergence acceleration. 

\subsection{Preconditioning for Fully Connected Networks}
\label{BNP_section}

We develop a preconditioning method for neural networks by considering gradient descent for parameters in one layer.  
In the theorem below, we derive the Hessian of the loss function with respect to a single weight vector and single bias entry. 
General formulas on gradients and the Hessian have been derived in \cite{Naumov17}. Our contribution here is to present the Hessian in a simple expression that demonstrates its relation to the mini-batch activations. 

Consider a neural network as defined in  (\ref{vanilla_eqn}); that is $h^{(\ell)} = g\left(W^{(\ell)}h^{(\ell-1)} + b^{(\ell)}\right)\in \mathbb{R}^{n_{\ell}}$. 
We denote $h_{i}^{(\ell)} = g\left(a_{i}^{(\ell)} \right)$ as the $i$th entry of $h^{(\ell)}$ where $a_{i}^{(\ell)} = w_{i}^{(\ell)^T}h^{(\ell-1)} + b_{i}^{(\ell)} \in \mathbb{R}$.  Here $w_{i}^{(\ell)^T} \in \mathbb{R}^{1 \times n_{\ell-1}}$ and $b_{i}^{(\ell)}$ are the respective $i$th row and entry of $W^{(\ell)}$ and $b^{(\ell)}$, and $n_{\ell-1}$ is the dimension of $h^{(\ell-1)}$.  Let 
\begin{equation}
\widehat{w}^{T} = \left[b_{i}^{(\ell)}, w_{i}^{(\ell)^T} \right] \in \mathbb{R}^{1 \times (n_{\ell-1}+1)},\;
\widehat{h} = \begin{bmatrix}
       1  \\
        h^{(\ell-1)}
      \end{bmatrix}
 \in \mathbb{R}^{(n_{\ell-1}+1) \times 1},\;
\mbox{ and }\; a_{i}^{(\ell)} = \widehat{w}^{T}\widehat{h}.
    \label{eq:ai} 
\end{equation}
 
\begin{proposition}
\label{grad_hessian_theorem}
Consider a loss function $L$  defined from the output of a fully connected multi-layer neural network for a single network input $x$. Consider the weight and bias parameters $w_{i}^{(\ell)}, b_{i}^{(\ell)}$ at the $\ell$-layer and 
write $L = L\left(a_{i}^{(\ell)} \right) = L\left( \widehat{w}^{T}\widehat{h} \right)$ as a function of the parameter $\widehat{w}$ through $a_{i}^{(\ell)}$ as in (\ref{eq:ai}).  
When training over a mini-batch of $N$ inputs,
let $\{h_{1}^{(\ell-1)}, h_{2}^{(\ell-1)}, \hdots, h_{N}^{(\ell-1)} \}$  be the associated $h^{(\ell-1)}$
and let $\widehat{h}_j=\begin{bmatrix}
       1  \\
        h_j^{(\ell-1)}
      \end{bmatrix}
 \in \mathbb{R}^{(n_{\ell-1}+1)\times 1}$. Let $$\mathcal{L} =\mathcal{L} (\widehat{w}) := \frac{1}{N}\sum_{j = 1}^{N}L\left(\widehat{w}^T\widehat{h}_j \right)$$ be the mean loss over the mini-batch. Then, its Hessian with respect to $\hat{w}$     is 
\begin{equation}
    \nabla^2_{\widehat{w}} \mathcal{L} (\widehat{w}) =  \widehat{H}^{T}S\widehat{H}  \label{hessian_eqn1} \end{equation}
    \text{where } $\widehat{H}=[e, \; H]$, 
    \begin{equation} {H} = \begin{bmatrix}
        h_{1}^{(\ell-1)^T} \\
       \vdots\\
        h_{N}^{(\ell-1)^T},
      \end{bmatrix} 
      \text{ and } 
      S=\frac{1}{N}\begin{bmatrix}
       L''\left(\widehat{w}^T\widehat{h}_1\right) & & \\
        &\ddots &\\
        & &L''\left(\widehat{w}^T\widehat{h}_N\right)
      \end{bmatrix}, 
    \label{hessian_eqn1part2} 
\end{equation}
with all off-diagonal elements of $S$ equal to 0.
%
\end{proposition}

Proposition \ref{grad_hessian_theorem} shows how the Hessian of the loss function relates to the neural network mini-batch activations. Since $\kappa(\nabla^2_{\widehat{w}} \mathcal{L}) \le \kappa( \widehat{H})^2 \kappa(S)$ (see Proposition 11 in Appendix), our goal is to improve the Hessian conditioning through that of $\widehat{H}$.
Although the matrix $S$ could also cause ill-conditioning, there is no good remedy. Thus we focus in this work on using the activation information to improve the condition number of $\widehat{H}$.

One common cause of ill-conditioning of a matrix is due to different scaling in its columns or rows. Here, if the features (i.e. the entries) in $h^{(\ell-1)}$  have different orders of magnitude, then the columns of  $\widehat{H}$ have different orders of magnitude  and $\widehat{H}$ will typically be ill-conditioned. 
Fortunately, this can be remedied by column scaling. The conditioning of a matrix may also be improved by making columns (or rows) orthogonal as is in an orthogonal matrix. 
Employing these ideas to improve conditioning of  $\widehat{H}$, 
we propose the following preconditioning transformation:  
\begin{equation}
\label{PD_eqn}
\widehat{w} = Pz, \;\;\mbox{ with }\;\;
P:=UD, \;\;
    U := \begin{bmatrix}
        1 & -\mu_H^{T}\\
        0 & I
        \end{bmatrix},\;\;\;
    D := \begin{bmatrix}
        1 & 0\\
        0 & \text{diag}\left(\sigma_H\right)
        \end{bmatrix}^{-1},
\end{equation}
where 
\begin{equation}
\label{mu}
\mu_H := \frac{1}{N} H^T e = \frac{1}{N}\sum_{j=1}^{N} {h}_j^{(\ell-1)}, \,\,\mbox{ and }\;\;
\sigma_H^2 :=  \frac{1}{N}\sum_{j=1}^{N} ({h}_j^{(\ell-1)}-\mu_H)^2
\end{equation}
are the (vector) mean and variance of $\{h_j^{(\ell-1)}\}$ respectively. 
Then, with this preconditioning, the corresponding Hessian matrix in $z$ is $P^T \nabla^2_{\widehat{w}} \mathcal{L} P= P^T   \widehat{H}^{T}S\widehat{H} P$ and
\[
\widehat{H} P = \begin{bmatrix}
     1 &(h_1^{(\ell-1)^T} -\mu_H^T) \\
      \vdots  &\vdots \\
      1  & (h_N^{(\ell-1)^T} -\mu_H^T)
      \end{bmatrix} D \\
      =\begin{bmatrix}
     1 &\frac{h_1^{(\ell-1)^T} -\mu_H^T}{\sigma_H} \\
      \vdots  &\vdots \\
      1  &\frac{h_N^{(\ell-1)^T} -\mu_H^T}{\sigma_H}
      \end{bmatrix}.
\]
Namely, the $U$ matrix centers  $\{h_j^{(\ell-1)}\}$ and then the scaling matrix $D$ normalizes the variance. We summarize this as the following theorem.

\begin{theorem}
\label{precond_hess}
For the loss function $\mathcal{L} =\mathcal{L} (\widehat{w}) $ defined in Proposition \ref{grad_hessian_theorem},
if we use the preconditioning transformation $\hat{w}=Pz$ in (\ref{PD_eqn}), then the preconditioned Hessian matrix is 
\[
\nabla^2_{z} \mathcal{L} =P^T \nabla^2_{\widehat{w}} \mathcal{L} P=\widehat{G}^T S \widehat{G}.
\]
where 
\begin{equation}
\widehat{G}= \begin{bmatrix}
       1 & g_{1}^T \\
       \vdots & \vdots\\
       1 & g_{N}^T
      \end{bmatrix} :=\widehat{H}P.
\label{Gmatrix}
\end{equation}
and $g_j= ({h}_j^{(\ell-1)}-\mu_H)/\sigma_H$ is ${h}_j^{(\ell-1)}$
normalized to have zero mean and unit variance.
\end{theorem}

In other words, the effect of preconditioning is that the corresponding Hessian matrix is generated by the normalized feature vectors $g_j$. 
Our next theorem shows that 
this normalization of ${h}_j^{(\ell-1)}$ improves the conditioning of $\nabla^2_{\widehat{w}} \mathcal{L}$ in two ways. 
First, note that 
$$
\widehat{H}U=\left[e,\; H - e \mu_H^T\right]
\;\;
\mbox{ and } \;\;
(H - e \mu_H^T)^T e=0.
$$ 
So multiplying $\widehat{H}$ by $U$ makes the first column orthogonal to the rest. This will be shown to improve the condition number. Second, $\sigma_H$ is the vector of the 2-norms of the columns of $H - e \mu_H^T$ scaled by $\sqrt{N}$. Thus, multiplying $H - e \mu_H^T$ by $D$ scales all columns of $H - e \mu_H^T$ to have the same norm of $\sqrt{N}$. This  also improves the condition number in general  by a theorem of  \cite{van1969condition} (given in Appendix A as Lemma A1; see also \citep[Theorem 7.5]{doi:10.1137/1.9780898718027}).We summarize in the following theorem. 


\begin{theorem}\label{thm:pcond} Let $\widehat{H}=[e, H]$ be the extended hidden variable matrix, $\widehat{G}$ the normalized hidden variable matrix, $U$ the centering transformation matrix, and $D$ the variance normalizing matrix defined by (\ref{hessian_eqn1}), (\ref{PD_eqn}), and (\ref{Gmatrix}). Assume $\widehat{H}$ has full column rank. We have
 \begin{equation*}
 \kappa(\widehat{H}U) \le   \kappa(\widehat{H}). \end{equation*}
This inequality is strict if $\mu_H\ne 0$ and is not orthogonal to $x_{\max}$, where $x_{\max}$ is an eigenvector corresponding to the largest eigenvalue of  the sample covariance matrix $\frac{1}{N-1}(H-e\mu_H^T)^T(H-e\mu_H^T)$ (i.e. principal component). Moreover, 
\begin{equation*}
 \kappa(\widehat{G}) =\kappa (\widehat{H} U D) \le \sqrt{n_{\ell-1}+1} \min_{D_0 \mbox{ is diagonal}} \kappa (\widehat{H} U D_0).
 \label{sluis_eqn}
 \end{equation*}
\end{theorem}

We remark that,  if $\mu_H=0$ (i.e. the data is already centered),  then $ U =I$ and hence no improvement in conditioning is made by $\widehat{H}U$. 
The second bound can be strengthened with the $\sqrt{n_{\ell-1}+1}$ factor removed, if $U^T\widehat{H}^T\widehat{H} U$ has a so-called {\em Property A} (i.e. there exists a permutation matrix $P$ such that $P(U^T\widehat{H}^T\widehat{H} U)P^T$ is a $2\times 2$ block matrix with the (1,1) and (2,2) blocks being diagonal); see \cite[p.126]{doi:10.1137/1.9780898718027}. In that case, the scaling by $D$ yields optimal condition number, i.e. $\kappa(\widehat{G}) =  \min_{D_0} \kappa (\widehat{H} U D_0)$.  However, Property A is not likely to hold in practice; so this result only serves to illustrate that the scaling by $D$ yields a nearly optimal condition number with  $\sqrt{n_{\ell-1}+1}$  being a pessimistic bound.

In general, if the entries of  $\sigma_H$ (or the diagonals of $D$) have different orders of magnitude, then the columns of $\widehat{H} U$ are badly scaled and is typically   ill-conditioned. This can be seen from  
\[
\kappa (\widehat{H} U) \le \kappa (\widehat{H} UD) \kappa(D^{-1}) = \kappa (\widehat{H} UD) \kappa(D) \le \kappa (D) \sqrt{n_{\ell-1}+1} \min_{D_0} \kappa (\widehat{H} U D_0).
\]
Other than the factor $\sqrt{n_{\ell-1}+1}$, each of the inequalities above is tight. So, $\kappa (\widehat{H}) \ge \kappa (\widehat{H} U) $ can be as large as $ \kappa (D) \min_{D_0} \kappa (\widehat{H} U D_0) $. 
In that situation, $\kappa(\widehat{G}) \le \sqrt{n_{\ell-1}+1} \min_{D_0} \kappa (\widehat{H} U D_0)$ may be as small as $ \sqrt{n_{\ell-1}+1} \kappa (\widehat{H}) / \kappa (D)$.
Thus, the preconditioned Hessian $\widehat{G}^T S \widehat{G}$ may improve the condition number of  $\widehat{H}^T S \widehat{H}$ by as much as  $ \kappa (D)^2$.

Over one training iteration, the improved conditioning as shown by Theorem \ref{thm:pcond} implies reduced $r$ and hence improved reduction of the error (\ref{eq:oneiter}) for that iteration. With mini-batch training, the loss function and the associated Hessian matrix change at each iteration, and so does the preconditioning matrix. Thus, our preconditioning transformation attempts to adapt to this change of the Hessian  and mitigates its effects. Globally, with the loss changing at each step, it is  difficult to quantify the degree of improvement in conditioning for all steps and how convergence improves over many steps as even the local minimizer changes. So our results only demonstrate the potential  beneficial effects of preconditioning at each iteration.

Since we use one learning rate for all the layer parameters, the convergence theory in Section \ref{pgd_section} shows that we need the learning rate to satisfy $\alpha < 
{2}/{\|\nabla^2_{\theta}L(\theta^{*}) \|}$ for the Hessians of all layers.
Then, a large norm of the Hessian  for one  particular layer will require a smaller learning rate overall.  It is thus desirable to scale all Hessian blocks 
to have similar norms. With the BNP transformation, the transformed Hessian is given by $\widehat{G}$ whose norm can be estimated using a result from random matrix theory as follows. 

First, we write $\widehat{G}=\widehat{H}UD=\left[e,\; G\right]\in \R^{N\times (n_{\ell-1}+1)}$, where $G=(H - e \mu_H^T) \text{diag}\left(\sigma_H\right)^{-1}$. Recall that  
$N$ is the mini-batch size. Since $e$ is orthogonal to $H - e \mu_H^T$ and hence to $G$, we have $\|\widehat{G} \|= \max\{\|e\|, \|G\|\} = \max\{\sqrt{N}, \|G\|\}$.  Furthermore, 
as a result of normalization,  the entries of the matrix $G=[g_{ij}]\in \R^{N\times n_{\ell-1}}$ satisfy $\sum_{i,j} g_{ij} =0$ and $\frac{1}{n_{\ell-1}N}\sum_{i,j} g_{ij}^2 =1$. We can therefore model the entries of $G=[g_{ij}]$  as random variables with zero mean and unit variance. In addition, the entries can be considered approximately independent. Specifically, if the input $x$ to the network has independent components and the weights and biases of the network are all independent, then the components of $h^{(\ell-1)}$  are also independent and so are their normalizations. Hence, for an iid mini-batch inputs, the entries of $G$ are also independent. Note that the weights and biases are typically initialized to be  iid although, after training, this is no longer the case.  If the independence  holds, then, using a theorem in random matrix theory \cite[Corollary 2.2]{seginer_2000} (see also Lemma A2 in Appendix), we have that the expected value of $\|G\|$ is bounded by some constant $C$ times $\max\{\sqrt{n_{\ell-1}}, \sqrt{N}\}$.  
Then the same holds for $\|\widehat{G} \| \le C \max\{\sqrt{n_{\ell-1}}, \sqrt{N}\}$.
Since $N$ is common for all layers, if we 
scale  $\widehat{G}$ by $1/q$ where $q^2 = {\max\{{n_{\ell-1}}/{N}, 1\}}$, then  $(1/q) \widehat{G}$ is bounded by $\sqrt{N}$, independent of $n_{\ell-1}$, the dimension of a hidden layer.  Hence, the Hessian blocks for different layers are expected to have comparable norms. This is summarized by the following proposition.
\begin{proposition} \label{randommatrix}
Let $n_{\ell-1}\ge 3$ and assume the entries of the normalized hidden variable matrix, $G\in \R^{N\times n_{\ell-1}}$ are iid  random variables with zero mean and unit variance. Then the expectation of the norm of  $(1/q) \widehat{G}= (1/q)\left[e,\; G\right]$ is bounded by $C \sqrt{N} $ for some constant $C$ independent of $n_{\ell-1}$, where $q^2 = {\max\{{n_{\ell-1}}/{N}, 1\}}$. 
\end{proposition}
We call the preconditioning by $(1/q)P$ a {\em batch normalization preconditioning} (BNP). 
For the $\ell$th layer, it takes the mini-batch activation $H = [h_{1}^{(\ell-1)}, h_{2}^{(\ell-1)}, \hdots, h_{N}^{(\ell-1)}]^T$ and the gradients  $\frac{\partial \mathcal{L}}{\partial W^{(\ell)}}\in  \mathbb{R}^{n \times n_{\ell-1}},\; \frac{\partial \mathcal{L}}{\partial b^{(\ell)}}\in  \mathbb{R}^{1 \times n}$ computed as usual and then modifies the gradient (see (\ref{pd_eqn3})) by the preconditioning transformation in the  gradient descent iteration as follows: 
    \begin{align}
     \label{eqn:train_step}
     \begin{bmatrix}
         b^{(\ell)^T}\\
         W^{(\ell)^T} 
     \end{bmatrix} \leftarrow 
     \begin{bmatrix}
         b^{(\ell)^T}\\
         W^{(\ell)^T}
     \end{bmatrix}
     - \alpha \frac{1}{q^2}PP^{T}
     \begin{bmatrix}
         \frac{\partial \mathcal{L}}{\partial b^{(\ell)}}\\
         \frac{\partial \mathcal{L}}{\partial W^{(\ell)^T}}
     \end{bmatrix}; \;\;
     P  = \begin{bmatrix}
     1 & -\mu^{T}\\
     0 & I
     \end{bmatrix}
     \begin{bmatrix}
     1 & 0\\
     0 & \text{diag}\left(\frac{1}{{\tilde{\sigma}}}\right)
     \end{bmatrix}.
 \end{align}   
We summarize the process as  Algorithm \ref{bnp_alg}.

\begin{algorithm}
        \caption{One Step of BNP Training on  $W^{(\ell)},  b^{(\ell)}$ of the $\ell$th Dense Layer}\label{bnp_alg}
        \begin{algorithmic}
        \label{alg1}
            \State \textbf{Given:}  $\epsilon_1=10^{-2}, \epsilon_2=10^{-4}$ and  $\rho=0.99$; learning rate $\alpha$;\\ initialization: $\mu = 0, \sigma = 1$;
            \State \textbf{Input:}  Mini-batch output of previous layer $H = [h_{1}^{(\ell-1)}, h_{2}^{(\ell-1)}, \hdots, h_{N}^{(\ell-1)}]^T \subset \mathbb{R}^{n_{\ell-1}}$
           and the  parameter gradients: $G_w\leftarrow \frac{\partial \mathcal{L}}{\partial W^{(\ell)}}\in  \mathbb{R}^{n_{\ell} \times n_{\ell-1}},\; G_b \leftarrow \frac{\partial \mathcal{L}}{\partial b^{(\ell)}}\in  \mathbb{R}^{1 \times n_{\ell}}$
            \State 
           1. Compute mini-batch mean/variance: $\mu_H, \sigma_H^2$; \\
            2. Compute running average statistics: $\mu \gets \rho\mu + (1-\rho)\mu_H$,  $\sigma^2 \gets \rho\sigma^2 + (1-\rho)\sigma_H^{2}$;

      \State     
      3. Set $\tilde{\sigma}^2 = \sigma^2 + \epsilon_1\max\{\sigma^2\} + \epsilon_2$ and $q^2 = \text{max}\{n_{\ell-1}/N, 1\}$;
     \State 
     4. Update $G_w$:  $G_w(i,j) \leftarrow\frac{1}{q^2}[G_w(i,j)-\mu(j)G_b(i)]/\tilde{\sigma}^2(j)$;
     \State 
     5. Update $G_b$:  $G_b(i) \leftarrow \frac{1}{q^2}G_b(i)-\sum_j G_w(i,j)\mu(j)$;

\State \textbf{Output:} Preconditioned gradients:  $G_w, G_b$.
    

        \end{algorithmic}
    \end{algorithm}

In Algorithm \ref{bnp_alg}, $\text{max} \{ \sigma^2\} $ denotes the maximum entry of the vector $\sigma^2 \in \mathbb{R}^{n_{\ell-1}}$ and $\tilde{\sigma}^2$ is ${\sigma}^2$ with a small number added to prevent division by a number smaller than $\epsilon_1\text{max} \{ \sigma^2\} $ or $\epsilon_2$. The $\mu$ and $\sigma$ in (\ref{eqn:train_step}) may be the mini-batch statistics $\mu_H$ and $\sigma_H$ or some averages of them. We find mean and variance  estimated using the running averages as in Step 2 works better in practice.  
In the case of mini-batch size 1, the mean is $\mu_H=h_1^{(\ell -1)}$ but the variance $\sigma_H$ is not meaningful. In that case, a natural generalization is to use the running mean $\mu$ to estimate the variance $\sigma_H^2=(h_1^{(\ell -1)}- \mu)^2$. Steps 4 and 5 implement the gradient transformation  in (\ref{eqn:train_step}), where the multiplications by $P$ and $P^T$ simplify to involve matrix additions and vector multiplications only.

Algorithm \ref{bnp_alg} has very modest computational overhead. The complexity to compute the preconditioned gradient (Step 4) is $6n_{\ell-1}n_{\ell}+2n_{\ell}$. Other computational costs of the algorithm consists of $4n_{\ell-1}N$ for the mini-batch mean and variance at Step 1, and $8n_{\ell-1}$ for $\mu$ and $\tilde \sigma$ at Step 2 and Step 3.  The total complexity for preconditioning is summarized as follows. 
 \begin{proposition}
 \label{flops}
  The number of floating point operations used in one step of BNP Training as outlined in Algorithm \ref{bnp_alg} is $4n_{\ell-1}N+6n_{\ell-1}n_{\ell}+2n_{\ell}+8n_{\ell-1}$.
 \end{proposition}

In comparison, BN also computes the mini-batch statistics $\sigma_H$ and $\mu_H$ and their moving averages, which amounts to $4n_{\ell-1}N+7n_{\ell-1}$ if BN is applied to $h^{(\ell-1)}$.  Additionally, BN  requires $4n_{\ell-1}N$ operations in the normalization step in 
the forward propagation including re-centering and re-scaling. For the back propagation through the BN layer, BN requires  $24 n_{\ell-1}N-3 n_{\ell-1}$ operations, including  the cost of passing gradients through $\sigma_H$ and $\mu_H$. 
The total cost of BN is $32n_{\ell-1}N+4n_{\ell-1}$.
Comparing with BNP's $n_{\ell-1}(4N+6n_{\ell})+2n_{\ell}+8n_{\ell-1}$ operations and ignoring first order terms, BN is more efficient if $28N<6n_{\ell}$ but more expensive otherwise. For a typical neural network implementation, we expect $28N>6n_{\ell}$ and so BNP has lower complexity. 

Finally, we note that the boundedness of the Hessian or flatness of the loss has been shown to have some relations to generalization (see \cite{DBLP:journals/corr/abs-1906-07774}), although \cite{DBLP:journals/corr/DinhPBB17} suggests this is not always the case. Our preconditioning uses an implicit transformation that improve the condition number of the Hessian for the transformed parameters. This only affects the gradients used in training; neither the network parameters nor the hidden activation are explicitly transformed. Namely,  the loss function stays the same. Whether the modified training algorithm could find a better minimizer is not clear. 

\subsection{Relation to BN}
\label{relation_section}

We discuss in this section the relation between BNP and BN. 
We first observe that a post-activation BN layer defined in (\ref{post_eqn}) with $\mathcal{B}_{\beta,\gamma}\left( \cdot \right)$ is equivalent to one normalized with $\mathcal{B}_{0,1}\left( \cdot \right)$ (that is no re-scaling and re-centering) with transformed weight and bias parameters. Namely, $h^{(\ell)} =g\left(W^{(\ell)} \mathcal{B}_{\beta,\gamma}(h^{(\ell-1)}) +b^{(\ell)}\right) = g\left(\widehat{W}\mathcal{B}_{0,1}\left(h^{(\ell-1)}\right) + \widehat{b}\right)$,
where $\widehat{W} = W^{(\ell)}\text{diag}(\gamma)$ and $\widehat{b} = W^{(\ell)}\beta + b^{(\ell)}$. So, $\mathcal{B}_{\beta,\gamma}\left( \cdot \right)$ is simply an over-parametrized version of $\mathcal{B}_{0,1}\left( \cdot \right)$.  Indeed, any change in the parameters $\{W^{(\ell)}, b^{(\ell)}, \beta,\gamma\}$ can theoretically be obtained from a corresponding change in  $\{\widehat{W}, \widehat{b}\}$, although their training iterations through gradient descent will be different  without any clear advantage or disadvantage for either.
From a theoretical standpoint, we may consider $\mathcal{B}_{0,1}\left( \cdot \right)$ only. Note that the discussion above applies to post-activation BN only. For pre-activation BN, \cite{Frankle21} recently shows that a trainable beta and gamma can improve a pre-activation ResNet accuracy by 0.5-2\%.

The BN transform $\mathcal{B}_{0,1}\left( \cdot \right)$ can be combined with the affine transform $W^{(\ell)} \mathcal{B}_{0,1}(h^{(\ell-1)}) +b^{(\ell)}$ to obtain a vanilla network with  transformed parameters $\{\widehat{W}, \widehat{b}\}$. Although the two networks are theoretically equivalent, the training of the two networks are different. The training of the BN network is through gradient descent in the original parameter $\{W^{(\ell)}, b^{(\ell)}\}$, while we can train on $\{\widehat{W}, \widehat{b}\}$ of the combined vanilla  network. We show in the next theorem that training the BN network on $\{W^{(\ell)}, b^{(\ell)}\}$ is equivalent to training on $\{\widehat{W}, \widehat{b}\}$ of the underlying combined vanilla network with the BNP transformation of parameter.

During training of BN, the mini-batch statistics $\sigma_H$ and $\mu_H$ are considered functions of the previous layer parameters and gradient computations pass through these functions \citep{Ioffe15}. In understanding the transformation aspect of BN, we assume in the following theorem $\sigma_H$ and $\mu_H$ are  independent of the previous layer trainable parameters; that is gradients do not pass through them. 

\begin{proposition}
\label{bnp_equiv_theorem}
A post-activation BN network defined in (\ref{post_eqn}) with $\mathcal{B}_{0,1}\left( \cdot \right)$ is equivalent to  a vanilla network (\ref{vanilla_eqn})  with parameter 
$\{\widehat{W}, \widehat{b}\}$ where $\widehat{W} = W^{(\ell)}\text{diag}\left(\frac{1}{\sigma_H}\right)$ and $\widehat{b} = b^{(\ell)} - W^{(\ell)}\text{diag}\left(\frac{\mu_H}{\sigma_H}\right)$.  Furthermore, one step of gradient descent training of BN with $\mathcal{B}_{0,1}\left(\cdot \right)$ in $\{W^{(\ell)}, b^{(\ell)}\}$ without passing the gradient through $\mu_H$, $\sigma_H$ is equivalent to one step of BNP training of the vanilla network (\ref{vanilla_eqn}) with parameter $\widehat{W}^T, \widehat{b}$.
\end{proposition}


Over one step of training, it follows from this equivalency and the theory of BNP that the BN transformation improves the conditioning of the Hessian and hence accelerates convergence, as compared with direct training on the underlying vanilla network.   However, further training steps of BN are not equivalent to BNP since the BN network changes (even without any parameter update) when the mini-batch changes, whereas the same underlying network is used in BNP.  Namely, for a new training step, a new mini-batch is introduced, which changes the architecture of the BN layers. The corresponding underlying  vanilla network has a new  $\widehat{W} = W^{(\ell)}\text{diag}\left(\frac{1}{\sigma_H}\right)$ and $\widehat{b} = b^{(\ell)} - W^{(\ell)}\text{diag}\left(\frac{\mu_H}{\sigma_H}\right)$. In this iteration, BN would be equivalent to BNP applied to the changed underlying network with new $\widehat{W}$ and $\widehat{b}$ as the parameters. 

We note that the equivalency result illustrates one local effect of BN and provides understanding of how the BN transformation may improve convergence. A complete version of BN with trainable $\beta, \gamma$ and passing gradients through the mini-batch statistics through $\mu_H$, $\sigma_H$ has additional benefits that can not be accounted for by the Hessian based theory. On the other hand, it also limits BN to using mini-batch statistics in its training network, and prevents BN from using averages statistics that is beneficial when using a small mini-batch size.

\subsection{Relation to LayerNorm and GroupNorm}
\label{gn_section}
Layer normalization (LN) and Group normalization (GN) are alternatives to BN that differ in the method of normalization. LN \citep{ba2016layer} normalizes all of the summed inputs to the neurons in a layer \citep{ba2016layer} rather than in a batch, and can be used in fully connected layers. GN can be used in convolution layers and groups the channels of a layer by a given groupsize and normalizes within each group by the mean and variance \citep{wu2018group}. Note GN becomes LN if we set the number of groups equal to 1 in convolution layers. 

LN and GN use one input at a time and do not rely on mini-batch inputs. So, they are not affected by small mini-batch size. Furthermore, LN and GN do not require different training and inference networks. 
They have the same weight and scale invariant properties as BN \citep{Summers2020Four}. However, their normalizations do not concern statistical distributions associated with mini-batches and can not address the ill-conditioning of the Hessian caused by mini-batch training, as BN and BNP do. Because of this, it may be advantageous to combine LN/GN with BN/BNP. For example, \cite{Summers2020Four} advocates combining GN and BN with small mini-batch sizes. In our experiments with ResNets, we also find using BNP on a GN network is beneficial.

\section{BNP for Convolutional Neural Network (CNN)}
\label{CNN_section}
  
In this section, we develop our preconditioning method for CNNs. Mathematically, we need to derive the Hessian of the loss with respect to one weight and bias, 
from which a preconditioner $P$ can be constructed as before. 
Consider a linear  convolutional layer  of a CNN that has a $3$-dimensional tensor $\mathbf{h}\in \mathbb{R}^{r \times s \times c}$ 
as input and  a $3$-dimensional tensor $\mathbf{a} \in \mathbb{R}^{r \times s \times c'}$ as output. Here we consider {\em same} convolution with zero padding that results in the same spatial dimension for output and input, but all discussions can be generalized to other ways of padding easily. Let $\mathbf{w} \in \mathbb{R}^{k \times k \times c\times c'}$ be a $4$-dimensional tensor kernel ($k$ is odd) and  $b =[b_i] \in \mathbb{R}^{c'}$ 
a bias vector that define the convolutional layer. 

Then, for a fixed output channel $d$, we have 
$$
\mathbf{a}( \cdot, \cdot, d)=\sum_{i=1}^{c} \text{Conv}\left( \mathbf{h}(\cdot , \cdot, i), \mathbf{w}(\cdot, \cdot, i,d) \right)+b_d, $$
where 
$$
\text{Conv}(h(\cdot , \cdot),w(\cdot , \cdot))(i,j)=\sum_{p,q =1}^k h\left(i- \frac{k+1}{2}+p, j- \frac{k+1}{2}+q\right) w(p,q),
$$ with any $h, w$ set to 0 if the indices are outside of their bounds (i.e. zero padding).
After a convolution layer, a nonlinear activation is typically applied to $\mathbf{a}$, and often a pooling layer as well, to produce the hidden variable of the next layer \citep[see][Chapter 9 for details]{Goodfellow-et-al-2016}.  These layers do not involve any trainable parameters.


We first present a linear function relating the layer output $\mathbf{a}$ to the input  $\mathbf{h}$  as well as   $\mathbf{w}$  and  ${b}$. For a tensor $\mathbf{t}$, we denote by $\text{vec}({\mathbf{t}})$
the vector reshaping of the tensor by indexing from the first dimension onward. For example, the first two elements of  $\text{vec}( {\mathbf{t}}(\cdot, \cdot, \cdot))$ are ${\mathbf{t}}(1, 1, 1)$ and ${\mathbf{t}}(2, 1, 1)$.

\begin{figure}[ht]
    \centering
\includegraphics[scale=.75]{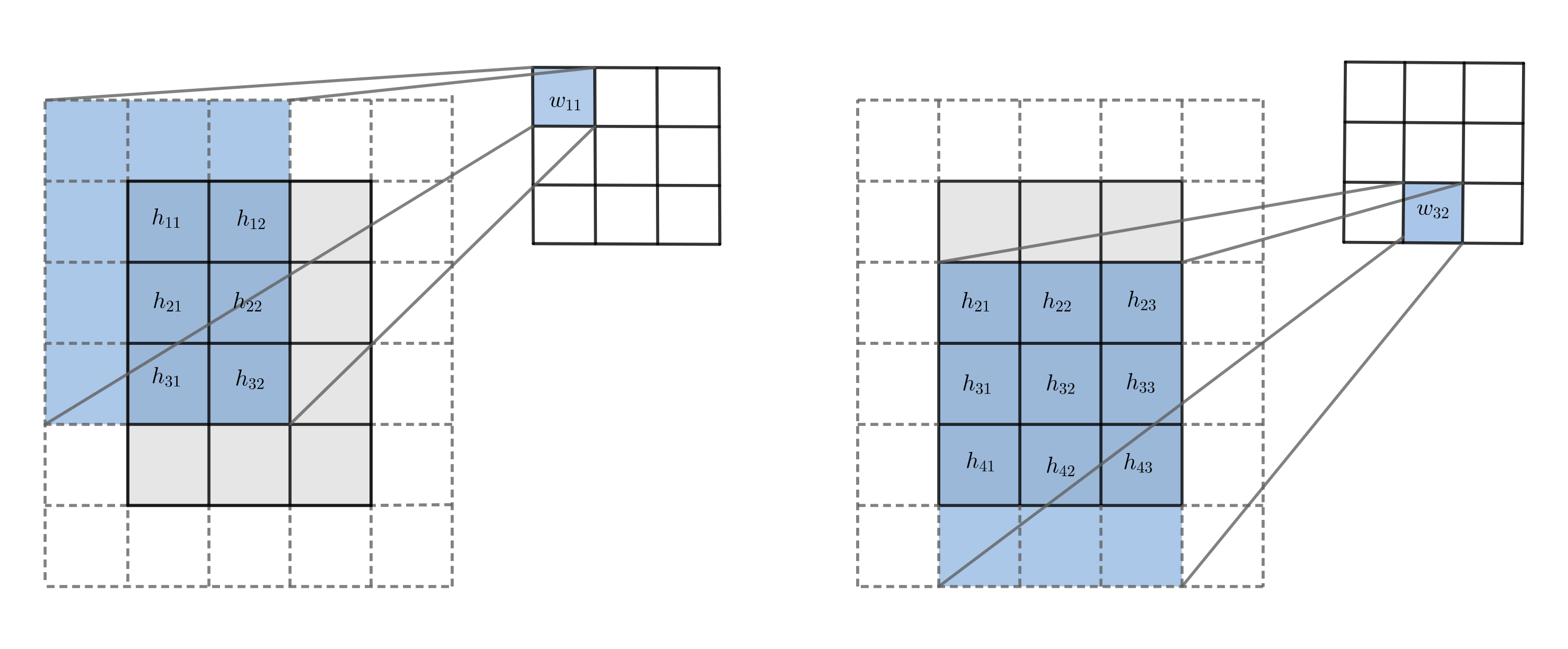}

    \caption{The blue shaded regions are $4\times 3$ submatrices in the $6\times 5$ zero padded  feature matrix traced out by a single element of the kernel during convolution. Illustrated here are the cases $w_{11}$ and $w_{32}$: the entries in the submatrix make up column 1 or column 6 of $\mathcal{H}$ which correspond to $w_{11}$ or $w_{32}$, respectively.}
\label{fig:Conv_diag}
\end{figure}

 Given $\mathbf{h}(\cdot, \cdot, p)$ and $\mathbf{w}(\cdot , \cdot, p,d)$, we can express the result of the 2-d convolution as a matrix vector product. To illustrate, first, consider for example that we have 
  $\mathbf{h}(\cdot, \cdot, p)=[h_{a,b}] \in \mathbb{R}^{4 \times 3}$ and $\mathbf{w}(\cdot, \cdot, p,d)=[w_{a,b}] \in \mathbb{R}^{3 \times 3}$  (omitting dependence on $p$ in $[h_{a,b}]$ and $ [w_{a,b}]$ notation below). Then, we can write $\text{vec(Conv}(\mathbf{h}(\cdot, \cdot, p),\mathbf{w}(\cdot, \cdot, p,d)))$ as
\\
\begin{center}
$\left[\begin{array}{ccc|ccc|ccc}
     &  &  &  & h_{11} & h_{21} & & h_{12} &  h_{22} \\ 
     &  &  &  h_{11}  &h_{21} & h_{31} & h_{12}& h_{22}& h_{32}\\
    & & & h_{21}& h_{31}& h_{41}  & h_{22}& h_{32} & h_{42} \\ 
        & & & h_{31}& h_{41}&   & h_{32}& h_{42} &  \\ \hline
    & h_{11} & h_{21} &  & h_{12}& h_{22} & & h_{13} & h_{23}\\
    h_{11} & h_{21} & h_{31} & h_{12} & h_{22} & h_{32} & h_{13}& h_{23} & h_{33} \\
    h_{21}  & h_{31} & h_{41}& h_{22} & h_{32} & h_{42} & h_{23} & h_{33}& h_{43}\\
        h_{31}  & h_{41} & & h_{32} & h_{42} &  & h_{33} & h_{43}&  \\\hline
    &h_{12} & h_{22}& & h_{13}  & h_{23} & &  & \\
    h_{12}       & h_{22} & h_{32} & h_{13} & h_{23}& h_{33} & &  & \\
        h_{22}   & h_{32} & h_{42} & h_{23} & h_{33} & h_{43} & &  &  \\
        h_{32}   & h_{42} &  & h_{33} & h_{43} &  & &  &  \\
\end{array}\right]
\cdot
\left[\begin{array}{c}
w_{11}\\
w_{21}\\
w_{31}\\
w_{12}\\
w_{22}\\
w_{32}\\
w_{13}\\
w_{23}\\
w_{33}\\
    \end{array}\right].$    
\end{center} 
Here, each column of this matrix corresponds to a kernel entry $w_{a,b}$ and consists of the elements of $\mathbf{h}(\cdot, \cdot, p)$ that are multiplied by that $w_{a,b}$ in the convolution. They are the elements of a submatrix of the padded $\mathbf{h}(\cdot, \cdot, p)$; see Figure \ref{fig:Conv_diag} for an illustration.

 In general, for fixed $p$ and $d$, we can write the 2-d convolution of $\mathbf{h}(\cdot, \cdot, p)$ and $\mathbf{w}(\cdot, \cdot, p,d)$ as a matrix vector product \begin{equation} \label{convolution_matrix_vector_product} \text{vec}(\text{Conv}(\mathbf{h}(\cdot, \cdot, p), \mathbf{w}(\cdot , \cdot, p,d)))=H_p \cdot \text{vec}(\mathbf{w}(\cdot, \cdot, p,d)),\end{equation}
 where $H_p \in \mathbb{R}^{(rs)\times (k^2)}$ and its $((i-1)r+ u, (j-1) k + v)$ entry is the entry of the padded feature map, $\mathbf{h}(\cdot , \cdot, p)$ that is multiplied by $\mathbf{w}(j, v, p,d)$ to obtain $\mathbf{a}(u,i,d)$ in the convolution, where $1 \leq u \leq r,$  $1 \leq i \leq s,$ and $1 \leq v,j \leq k$.
 As in the example above, this matrix can be written as a $s\times k$ block matrix of size $r \times k$, where the 
 $(t, u)$ block ($1 \leq t \leq s$, $1 \leq u \leq k$) of $H_p$ is 
\begin{align*}
    (H_p)_{tu}=\left[\begin{array}{cccc}
     \mathbf{h}(t',  u',p)& \mathbf{h}(t'+1, u',p)  & \hdots & \mathbf{h}(t'+(k-1), u',p)  \\ 
     \mathbf{h}(t'+1, u',p)& \mathbf{h}(t'+2, u',p) & \hdots & \mathbf{h}(t'+k, u',p) \\
    \vdots& \vdots &  & \vdots \\ 
    \mathbf{h}(t'+(r-1), u',p)& \mathbf{h}(t'+r, u',p) & \hdots & \mathbf{h}(t'+(r+k-2), u',p)\\
\end{array}\right], \end{align*} \\
where $t'=1-\frac{k-1}{2}$,  $u'=u-\frac{k-1}{2}+(t-1)$, and $\mathbf{h}(a,b,p)=0 $ for any index outside its bound. In particular, the $(j-1) k + v$-th column of $H_p$ consists of the entries $\mathbf{h}(u , i, p)$ in the $r\times s$ submatrix of the padded feature map where the submatrix is given by $u$ and $i$ in the ranges  $-\frac{k-1}{2}+1+(v-1)\leq u \leq r-\frac{k-1}{2}+(v-1)$ and $-\frac{k-1}{2}+1+(j-1)\leq i \leq s-\frac{k-1}{2}+(j-1)$.

Equation (\ref{convolution_matrix_vector_product}) allows us to relate the layer output to the input as 
\begin{align*}
\text{vec}(\mathbf{a}(\cdot, \cdot, d))&=\sum_{p=1}^c \text{vec}( \text{Conv}(\mathbf{h}(\cdot, \cdot, p), \mathbf{w}(\cdot , \cdot, p,d)))+b_d e \\
&=     \sum_{p=1}^c H_p \cdot \text{vec}(\mathbf{w}(\cdot, \cdot, p,d))+b_d e \\
&=\left[\begin{array}{ccccc} e & H_1 & H_2 & \hdots & H_c \end{array} \right]\cdot \widehat{w},
\end{align*}
where 
\begin{equation}\widehat{w}=\left[\begin{array}{c}
b_d\\
\omega_d
\end{array}\right]\label{w_hat} \in \R^{ck^2+1}
\;\mbox{ and }\;
\omega_d=
\left[\begin{array}{c} 
\text{vec}\left(\mathbf{w}(\cdot, \cdot, 1, d)\right)\\
\vdots \\
\text{vec}\left(\mathbf{w}(\cdot, \cdot, c, d)\right)
\end{array}\right] \in \R^{ck^2}.\end{equation}
Note that we use $\omega_d$ to denote the vector reshaping of the tensor  $\mathbf{w}(\cdot, \cdot, \cdot, d)$ for a fixed $d$ and, for simplicity, we drop the dependence on $d$ in the notation of $\widehat{w}$  as $d$ is fixed for all related discussions. 
When training over a mini-batch of $N$ inputs, we have one $\mathbf{a}$ for each input.
Combining the equations for all $N$ inputs in the mini-batch gives the linear relationship as summarized in the following lemma.


\begin{lemma}\label{CNN_notation_theorem} Given a mini-batch of $N$ inputs, let $\{\mathbf{h}^{(1)}, \cdots, \mathbf{h}^{(N)}\}$ be the associated input tensors at layer $\ell$, and $\{\mathbf{a}^{(1)}, \cdots,\mathbf{a}^{(N)} \}$ the output tensors. For a fixed output channel $d$, we have 
$\widehat{a}=\widehat{\mathcal{H}} \cdot \widehat{w},$ where 
$\widehat{w}$ is defined as in (\ref{w_hat}) to be the vectorized bias and weight elements, 
\begin{equation}\label{eq:Hhat}
{\small 
\widehat{\mathcal{H}} = [e, \mathcal{H}],\;\;
\mathcal{H}=
\left[\begin{array}{ccccc}
     H_1^{(1)}  & \hdots & H_c^{(1)}  \\ 
     H_1^{(2)} & \hdots & H_c^{(2)} \\
     \vdots&  & \vdots \\ 
     H_1^{(N)}  & \hdots & H_c^{(N)}\\
\end{array}\right],  \, 
\widehat{a}=
\left[\begin{array}{c}
\text{vec}\left(\mathbf{a}^{(1)}(\cdot, \cdot, d) \right)\\
\vdots\\
\text{vec}\left(\mathbf{a}^{(N)}(\cdot, \cdot, d) \right)\\
\end{array}\right], 
}
\end{equation}
and $H_p^{(i)} \in\R^{rs \times k^2}$ is as defined in  (\ref{convolution_matrix_vector_product}) from $\mathbf{h}^{(i)}$.
\end{lemma}

In the lemma, note that $\widehat{a}\in \R^{Nrs}$ and $\mathcal{H}\in \R^{Nrs \times k^2c}$.
With this linear relationship we derive the corresponding Hessian of the loss function. 

\begin{proposition} \label{CNN_hessian_theorem} 
Consider a CNN loss function $L$ defined for a single input tensor and write   ${L} ={L}(\mathbf{a}( \cdot, \cdot, d))$  as a function of the convolution kernel $\mathbf{w}( \cdot, \cdot, \cdot, d)$ through  $$\mathbf{a}( \cdot, \cdot, d){=\sum_{c=1}^{c_l} \text{Conv} (\mathbf{h}\left(\cdot , \cdot, c),  \mathbf{w}(\cdot, \cdot, c,d)\right)}+b_d.$$ When training over a mini-batch with $N$ inputs, let the associated hidden variables 
$\mathbf{h}$ of layer $\ell$ be  $\{ \mathbf{h}^{(1)}, \mathbf{h}^{(2)}, ...,\mathbf{h}^{(N)}\}$ 
and let the associated output of layer $\ell$ be  $\{ \mathbf{a}^{(1)}, \mathbf{a}^{(2)}, ...,\mathbf{a}^{(N)}\}$.
Let $\mathcal{L}  =\frac{1}{N}\sum_{j=1}^N L(\mathbf{a}^{(j)}( \cdot, \cdot, d))$ be the mean loss over the mini-batch. Then
\begin{equation*}
  \nabla_{\widehat{w}}^2 \mathcal{L}=  \mathcal{\widehat{H}}^TS\mathcal{\widehat{H}}, \end{equation*}
where $\mathcal{\widehat{H}}$ is as defined in (\ref{eq:Hhat}), $S=\frac{1}{N} \text{diag}\left\{
     \frac{\partial^2 L}{\partial v_a^2}(v_a^{(1)}), \frac{\partial^2 L}{\partial v_a^2}(v_a^{(2)}),
     \cdots, \frac{\partial^2 L}{\partial
    v_a^2}(v_a^{(N)})\right\}$, 
    $ v_a:=\text{vec}(\mathbf{a}( \cdot, \cdot, d))$, and $ v_a^{(j)}:=\text{vec}\left(\mathbf{a}^{(j)}( \cdot, \cdot, d)\right)$.  
 
       \end{proposition}

This Hessian matrix has a similar form as in the fully connected network (\ref{hessian_eqn1}).
We can apply preconditioning transformation (\ref{PD_eqn}) to improve the conditioning of $\mathcal{\widehat{H}}$ and hence of $\nabla_{\widehat{w}}^2 \mathcal{L}$ in the same way. As in the fully connected network, we use the preconditioner $P=UD,$ where 
\[
U=\begin{bmatrix}
     1& -\mu_H^T  \\ 
     0&  I \\
     
\end{bmatrix}, \, \,  
D=\begin{bmatrix}
     1& 0  \\ 
     0&  \text{diag}(\sigma_H)  
\end{bmatrix}^{-1},
\]
$\mu_H$ and $\sigma_H$ are defined from $\mathcal{H}$ as
\begin{equation*}\label{eq:cnnmean}
\mu_H:=\frac{1}{Nrs} {\mathcal{H}}^T e = \frac{1}{Nrs} \sum_{i=1}^{Nrs} \mathcal{H}_i, \,\;
\sigma_H^2:=\frac{1}{Nrs} \sum_{i=1}^{Nrs} (\mathcal{H}_i - \mu_H)^2,
\end{equation*}
and $\mathcal{H}^T_i$ is the $i$th row of $ \mathcal{H}$.
With this $P$, $\mathcal{\widehat{H}}P$ has the same property as ${\widehat{H}}P$ for the fully connected network. By applying Theorem \ref{thm:pcond} to $\mathcal{\widehat{H}}P$, the conditioning of $\mathcal{\widehat{H}}$ is improved with the preconditioning by $P$. 
    
Thus, the BNP transformation for a convolutional layer has the same form and the same property as in the fully connected network, except that the $\mu_H$ and $\sigma_H$ are the means and variances over the columns of $\mathcal{H}$. Noting that each column of $\mathcal{H}$ contains the elements of the feature maps that are multiplied by the corresponding kernel entry during the convolution for all inputs in a mini-batch, 
thus the means and variances are really the means and variances over the $N$ submatrices of feature maps defined by the kernel convolution for the mini-batch. See Figure \ref{fig:Conv_diag} for a depiction. Mathematically, we can express the $(p-1)k^2+(a-1)k+t$ entry of vector mean $\mu_H$ and vector variance $\sigma_H$  corresponding to ${\bf w}(t,a,p,d)$, indexed by $(t,a,p)$, as 
\begin{equation} \label{eq:CNN_mean_submatrix} \mu_H(t,a,p)=\frac{1}{Nrs}\sum_{i=1}^{N}\sum_{u=t-\frac{k-1}{2}}^{r+t-\frac{k-1}{2}-1}\sum_{v=a-\frac{k-1}{2}}^{s+a-\frac{k-1}{2}-1} h^{(i)}(u,v,p)\end{equation}
and 

\begin{equation} \label{eq:CNN_variance_submatrix} \sigma_H^2(t,a,p)=\frac{1}{Nrs}\sum_{i=1}^{N}\sum_{u=t-\frac{k-1}{2}}^{r+t-\frac{k-1}{2}-1}\sum_{v=a-\frac{k-1}{2}}^{s+a-\frac{k-1}{2}-1} (h^{(i)}(u,v,p)-\mu(t,a,p))^2, \end{equation}
for $1 \leq t,a \leq k$ and $1\le p \le c$. 

The entries in $\mu_H$ and $\sigma^2_H$ vary slightly from entry to entry, being the means and variances of  different submatrices of the zero-padded feature maps used, which  mostly overlap with the feature map. If $r,s$ are large,  they are all approximately equal to 
the mean and variance of the feature map $\mathbf{h} (\cdot, \cdot, p)$ over the  mini-batch  for each fixed input channel $p$, written as
\begin{equation}\label{eq:CNN_approx_mean}\displaystyle{\mu (p)=\frac{1}{Nrs}\sum_{i=1}^{N}\sum_{u=1}^{r}\sum_{v=1}^{s} h^{(i)}(u,v,p)}\end{equation}
and
\begin{equation}\label{eq:CNN_approx_var}\displaystyle{\sigma(p)^2=\frac{1}{Nrs}\sum_{i=1}^{N}\sum_{u=1}^{r}\sum_{v=1}^{s} (h^{(i)}(u,v,p)-\mu)^2},
\end{equation}
where $\mu$ and $\sigma$ are vectors with $\mu(p)$ and $\sigma(p)$ as entries.
Using this approximation significantly saves computations, which is adopted in our implementation of BNP. Indeed, this is how BN normalizes the activation of a convolution layer in CNNs. 

\begin{figure}[ht]
    \centering
\includegraphics[scale=.4]{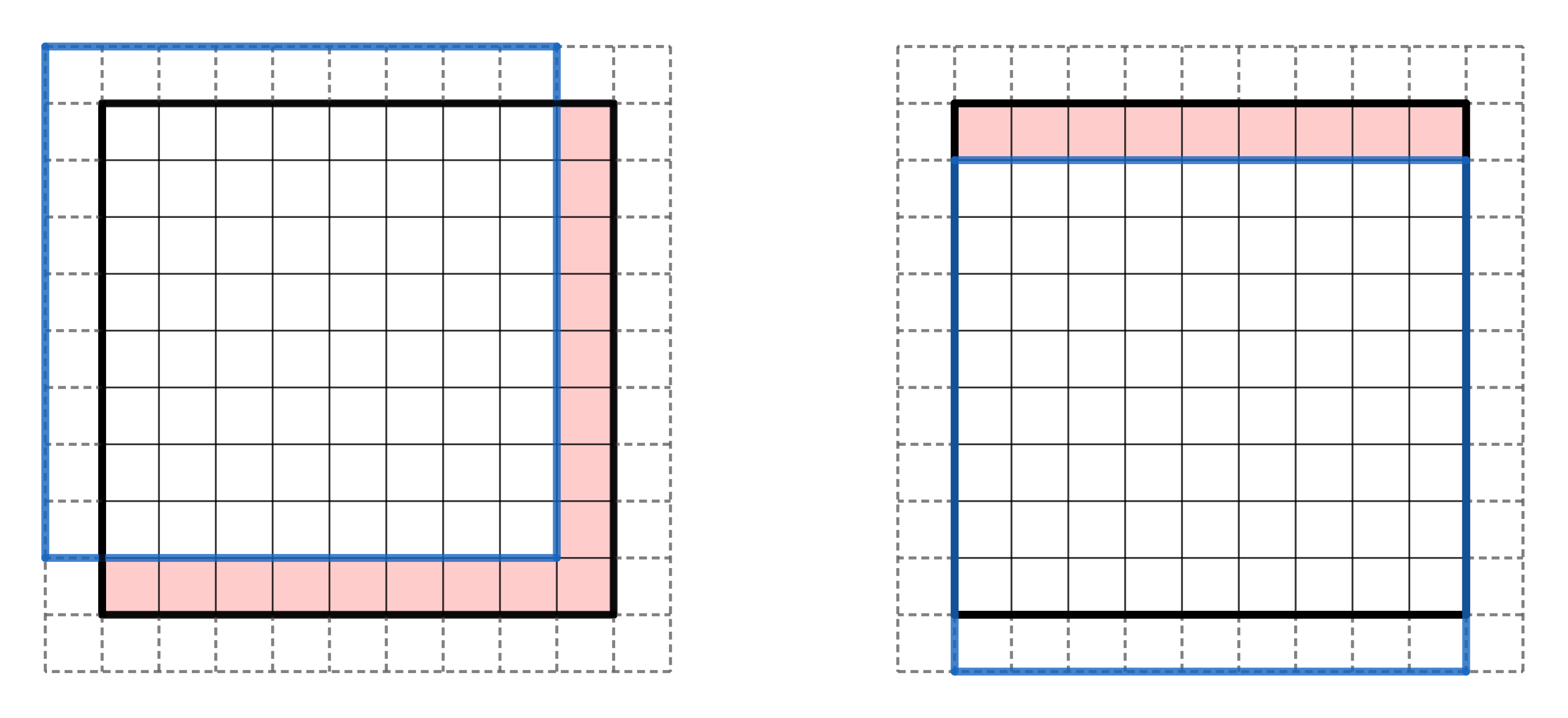}
    \caption{The  $9 \times 9$ blue submatrix of the zero padded feature map consists of the elements traced out by  $w_{11}$ (Left) or $w_{32}$ (Right) in the $3 \times 3$ kernel convolution. The mean $\mu_H$ and variance $\sigma_H$ are taken over these blue submatrices, while $\mu$ and $\sigma$ are taken over the $9 \times 9$ feature map, outlined in bold. The error in our approximation is given by the difference in elements, in shaded red above; those elements of the feature map excluded from the submatrix. There is a difference of at most $\frac{k-1}{2}$ rows and columns on the border of the feature map. Thus for $r,s \gg k$, this error is small.  }
\label{fig:Conv_error_diag}
\end{figure}

The error in approximating $\mu_H(i , j, p)$ and $\sigma_H(i,j,p)$  for all $i, j$ by $\mu(p)$ and $\sigma(p)$ is given by the difference between the feature maps and the submatrices of the padded feature maps. They have the same dimension  and differ at most in $\frac{k-1}{2}$ columns and rows on the border of the feature map, replacing these border entries with the zeros of the padded feature map; see Figure \ref{fig:Conv_error_diag} for an illustration. The error can be bounded as in the following proposition, which shows that, if the feature map size, $r,s$, is much bigger than the kernel size $k$, the error is small.


\begin{proposition} \label{CNN_error_bound_proposition}
Under the notation defined in Proposition \ref{CNN_hessian_theorem}, the error $|\mu(t,a,p)-\mu (p)|$ of approximating $\mu(t,a,p)$, the mean calculated for the BNP transformation, as defined in (\ref{eq:CNN_mean_submatrix}), by $\mu (p)$, as in (\ref{eq:CNN_approx_mean}), the mean over the entire feature map, is bounded by 
$$\displaystyle{(k-1)\left(\frac{1}{2r}+ \frac{1}{2s} -\frac{k-1}{4rs}\right)\max_{u,v,i}{|h^{(i)}(u,v,p)|}}.$$ 
\end{proposition}
As in Section \ref{BNP_section}, we further consider scaling each Hessian block to have comparable norms by estimating the norm of  $ \mathcal{H}$.  A direct application of Proposition \ref{randommatrix} to $\mathcal{\widehat{H}}P$ would indicate a norm bounded by  $\max\{\sqrt{k^2c}, \sqrt{Nrs}\}$, where $k^2c$ is the number of its columns and $Nrs$ is the number of its rows. But  one difficulty  is that $ \mathcal{H}$ contains many shared entries, so the entries of $\mathcal{\widehat{H}}P$ can no longer be modeled as independent random variables. We note that $ \mathcal{H}$ in (\ref{eq:Hhat}) has $N$ block rows where each block entry $H_p^{(i)}$ has $rs$ rows, but its columns are rearrangements of mostly the same entries. Namely, the entire matrix  $H_p^{(i)}$ contains only $rs$ independent entries. Thus, it may be effectively considered a matrix of independent entries 
with $\sqrt{rs}$ rows. Then $\mathcal{\widehat{H}}P$ may be regarded as having effectively $N\sqrt{rs}$ rows. We therefore suggest to use  $\max\{\sqrt{k^2c}, \sqrt{N}(rs)^{1/4}\}$ as a norm estimate. 
Thus, we scale $\mathcal{\widehat{H}}P$ by $q$, i.e. use $(1/q)P$ as a preconditioner, where   $q^2=\text{max}\{{k^2c/N}, \sqrt{rs}\}$. Our experiments with several variations of CNNs shows that this heuristic based estimate works well.


The detailed BNP preconditioning transformation for the $\ell$th convolution layer with the wight tensor ${\bf w}$ and bias vector $b$ (with their dependence on $\ell$ dropped in the notation for simplicity) is carried out in the vector reshaped from the tensor. 
Let $W=\begin{bmatrix}
         \omega_1, ..., \, \omega_{c_{\ell}}
     \end{bmatrix} \in \mathbb{R}^{k^2c_{\ell -1} \times c_{\ell}}$ be the matrix representation of ${\bf w}$ where the first three dimensions are reshaped into a vector, i.e.  $\omega_{d}=vec({\bf w}(\cdot, \cdot, \cdot, d)) $ (see (\ref{w_hat})). 
Let $\mu \in \mathbb{R}^{c_{\ell-1}}$ and $\sigma\in \R^{c_{\ell-1}}$ be defined in (\ref{eq:CNN_approx_mean}) and (\ref{eq:CNN_approx_var}), e.g. $\mu  =\begin{bmatrix}
          \mu(1),  \, .\, . \, ., \,  \mu(c_{\ell -1})
     \end{bmatrix}^T \in \mathbb{R}^{c_{\ell-1}}$. 
Then $\mu_H$ and $\sigma_H$ are approximated by  $\mu \otimes e  := \begin{bmatrix}
          \mu(1) e^T,  \, .\, . \, ., \,  \mu(c_{\ell -1}) e^T
     \end{bmatrix}^T \in \mathbb{R}^{k^2c_{\ell-1}}$ and $\sigma \otimes e  :=  \begin{bmatrix}
          \sigma(1) e^T,  \, .\, . \, ., \,  \sigma(c_{\ell -1}) e^T
     \end{bmatrix}^T \in \mathbb{R}^{k^2c_{\ell-1}}$, and $e =[1, 1, \cdots, 1]^T \in \mathbb{R}^{k^2}$. Then, the preconditioned gradient descent is the following update in $W$ and $b$: 
     \begin{align} \label{cnn_pd2pt}
            \begin{bmatrix}
         b^T\\
         W
     \end{bmatrix} \leftarrow 
     \begin{bmatrix}
         b^T\\
         W
     \end{bmatrix}
     - \alpha  \frac{1}{q^2}PP^{T}
        \begin{bmatrix}
         \frac{\partial \mathcal{L}}{\partial b}\\
          \frac{\partial \mathcal{L}}{\partial W}
     \end{bmatrix}; \;\;
     P  = \begin{bmatrix}
     1 & -\mu^T \otimes e^T\\
     0 & I
     \end{bmatrix}
     \begin{bmatrix}
     1 & 0\\
     0 & \text{diag}\left(\frac{1}{{\tilde{\sigma} \otimes e}}\right)
     \end{bmatrix},
    \end{align}
 
This vector form of the gradient descent update can be stated in the original tensor ${\bf w}$ with the matrix multiplications by $P$ and $P^T$ simplified. We present it in Algorithm \ref{cnn_bnp_alg}.

\begin{algorithm}
        \caption{One Step of BNP Training of a Convolution Layer with weight ${\bf w}$ and bias $b$ }\label{cnn_bnp_alg}
        \begin{algorithmic}
        \label{alg2}
            \State \textbf{Given:}  $\epsilon_1=10^{-2}, \epsilon_2=10^{-4}$ and  $\rho=0.99$; learning rate $\alpha$;\\ initialization: $\mu = 0, \sigma = 1$;
            \State \textbf{Input:}  Mini-batch output of previous layer $H = [h_{1}^{(\ell-1)}, h_{2}^{(\ell-1)}, \hdots, h_{N}^{(\ell-1)}]^T \subset \mathbb{R}^{ c_{\ell-1} \times r \times s}$
           and the  parameter gradients: $G_\mathbf{w}\leftarrow \frac{\partial \mathcal{L}}{\partial \mathbf{w}}\in  \mathbb{R}^{k \times k \times c_{ \ell-1} \times c_{\ell}} ; G_b \leftarrow \frac{\partial \mathcal{L}}{\partial b} \in  \mathbb{R}^{1 \times c_{\ell}}$
            \State 
           1. Compute mini-batch mean/variance: $\hat{\mu}, \hat{\sigma}^2 \in \mathbb{R}^{c_{\ell-1}}$ according to (\ref{eq:CNN_approx_mean}) and (\ref{eq:CNN_approx_var}); \\
            2. Compute running statistics: $\mu \gets \rho\mu + (1-\rho)\hat{\mu}$, $\sigma^2 \gets \rho\sigma^2 + (1-\rho)\hat{\sigma}^{2}$;
            
      \State     
      3. Set $\tilde{\sigma}^2 = \sigma^2 + \epsilon_1\max\{\sigma^2\} + \epsilon_2$ and $q^2 = \text{max}\{c_{\ell-1}k^2/N, \sqrt{rs}\}$;
     \State 
     4. Update $G_\mathbf{w}$: 
     $G_\mathbf{w}(i,j,p,d)\leftarrow \frac{1}{q^2}\left(G_\mathbf{w}(i,j,p,d)-\mu(p) G_b(d)\right)/ \tilde{\sigma}(p)^2$; 
     \State 
     5. Update $G_b$: $G_b(d)\leftarrow 
     \frac{1}{q^2}G_b(d) -\sum_{i,j,p} \mu(p) G_\mathbf{w}(i,j,p,d)$;  
     \State 
     {\bf Output:} updated gradients: $G_\mathbf{w}, G_b$
        \end{algorithmic}
    \end{algorithm}
    
Finally, we remark that our derivation of preconditioning $\mathcal{\widehat{H}}$ leads to the normalization by mean and variance over  the mini-batch and the two spatial dimensions. This not only provides a  justification to the conventional approach in applying BN to CNNs but also validates our theory. Note that a straightforward application of BN to CNNs may suggest  normalizing each pixel 
by the pixel mean and variance  over a mini-batch only. Under our theory, it is the Hessian matrix of Proposition \ref{CNN_hessian_theorem} that determines what mean and variance should be used in normalization. In particular, the means and variances of the columns of $\mathcal{\widehat{H}}$, the extended hidden variable matrix, naturally lead to the corresponding definitions.

Another interesting implication of our theory concerns online learning with mini-batch size $N=1$. Unlike BN, BNP is not constrained on the mini-batch size.  For the fully connected network, however, $\widehat{H}$ in (\ref{hessian_eqn1part2}) has 1 row only and hence a condition number of 1, which can not be reduced. So, although BNP may still be beneficial due to Hessian block norm balancing, the effect is not expected to be as significant as with larger $N$. However, for CNNs, $\mathcal{\widehat{H}}$ has dimension $Nrs \times (k^2 c+1)$ and even when $N=1$ the preconditioning can still offer improvements by Theorem \ref{thm:pcond}.

\section{Experiments}
\label{exp_main}
In this section, we compare BNP with several baseline methods on several architectures for image classification tasks. We also present some exploratory experiments to study computational timing comparison as well as improved condition numbers.

First, we present experiments comparing BNP with BN (and Batch Renorm when a small mini-batch size is used) and vanilla networks. Where suitable, we also compare with LayerNorm (LN) and GroupNorm (GN) as well as a version of BN with 4 things/improvements \citep{Summers2020Four}. We test them in three architectures for image classification: fully connected networks, CNNs, and ResNets.
All models use ReLU nonlinearities and the cross-entropy loss. Each model is tuned with respect to the learning rate. Default hyperparameters for BN and optimizers as implemented in Tensorflow or PyTorch are used, as appropriate. For BNP, the default values $\epsilon_1 = 10^{-2}$, $\epsilon_2 = 10^{-4}$, and $\rho=0.99$ are also used. Other detailed experimental settings are given in Appendix B.

{\bf Data sets:} We use MNIST, CIFAR10, CIFAR100, and ImageNet data sets. The MNIST data set \citep{LeCun13} consists of 70,000 black and white images of handwritten digits ranging from 0 to 9.  Each image is 28 by 28 pixels.  There are 60,000 training images and 10,000 testing images.  The CIFAR10 and CIFAR100 data sets \citep{CIFAR10} consist of 60,000 color images of 32 by 32 pixels with 50,000 training images and 10,000 testing images.  The CIFAR10 and CIFAR100 data sets consist of 10 and 100 classes respectively. The ImageNet datatset \citep{ILSVRC15} consists of 1,431,167 color images with 1,281,167 training images, 50,000 validation images, and 100,000 testing images. Each image has 256 by 256 pixels and the ImageNet dataset consists of 1,000 classes.



{\bf Fully Connected Neural Network.} We follow  \cite{Ioffe15} and consider a fully connected network consisting of three hidden layers of size 100 each and an output layer of size 10.  
We test it on the MNIST and CIFAR10 data sets.
For each experiment, we flatten each image into one large vector as input.  For the BN networks, normalization is applied post-activation. For this experiment, we also compare against an additional baseline of LayerNorm (LN), which is not affected by small batch size. 
Each model is trained using SGD.
\begin{figure}[ht]

    \centering
    \begin{subfigure}{.49\textwidth}
    \centering
    \includegraphics[width=1\columnwidth]{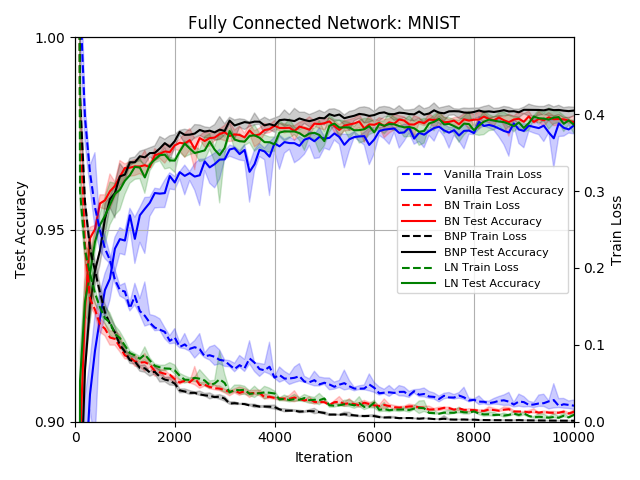}
    \end{subfigure}
    \begin{subfigure}{.49\textwidth}
    \centering
    \includegraphics[width=1\columnwidth]{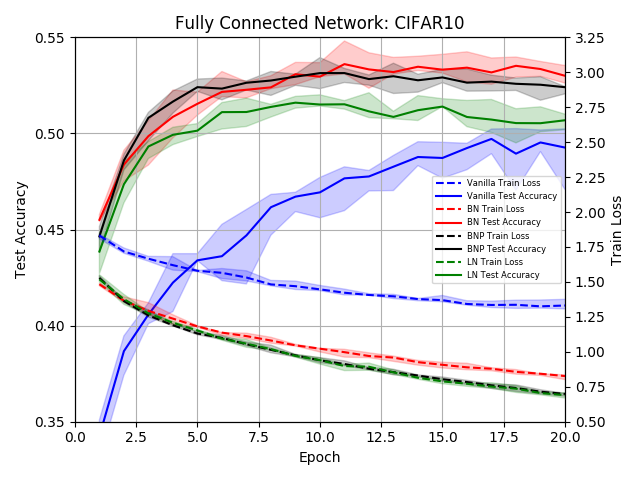}
    \end{subfigure}
    
    \caption{Fully connected network with mini-batch size 60: Training loss (dashed lines) and test accuracy (solid lines) for the vanilla network, BN, LN, and BNP on MNIST (Left) and CIFAR10 (Right).  The lines graph the means while the shaded regions graph the ranges of 5 tests.
    }
   \label{fig:dense_60}
\end{figure}


We first test on MNIST and CIFAR10 with mini-batch size 60, and the results, averaged over 5 runs with the mean and the range, are plotted  in Figure \ref{fig:dense_60}. BN and BNP have comparable performance in the MNIST experiments, with BN slightly faster at the beginning and BNP outperforming after that. They slightly outperform LN and all improve over  the vanila network. 
 For the CIFAR10 experiment, we see that BNP and BN are also comparible, with BNP converging slightly faster at the beginning but mildly overfitting after the 10th epoch, while BN achieving slightly better final accuracy without any overfitting. 
In both cases, BNP and BN  outperform the vanilla and LN networks. In general, BN does well in large mini-batch sizes and is less likely to overfit. 

\begin{figure}[ht]
    \centering
    \begin{subfigure}{.49\textwidth}
    \centering
    \includegraphics[width=1\columnwidth]{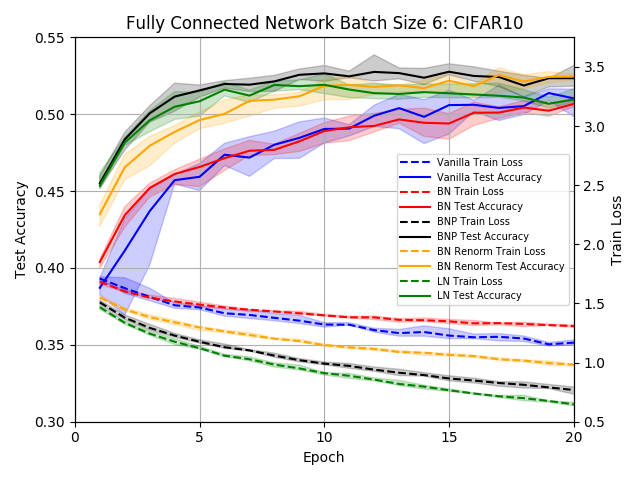}
    \end{subfigure}
    \begin{subfigure}{.49\textwidth}
    \centering
    \includegraphics[width=1\columnwidth]{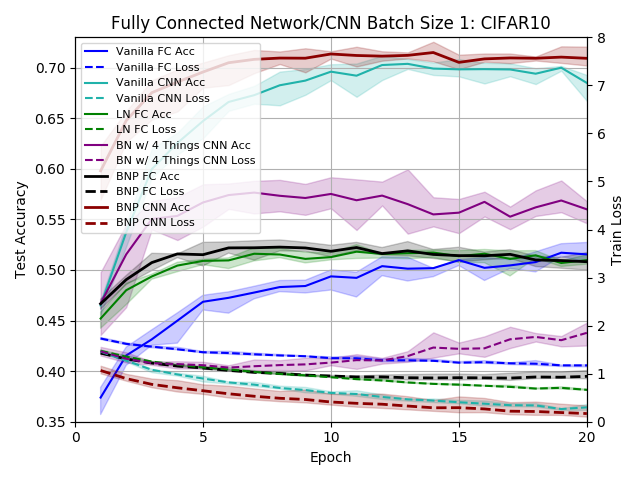}

    \end{subfigure}
    
    \caption{Fully connected (FC) network/CNN with small mini-batch sizes (6 and 1): Training loss (dashed lines) and test accuracy (solid lines). Left: fully connected network with mini-batch size 6 for vanilla network, BN, BN Renorm, LN, and BNP; Right: fully connected (FC)  network and 5-layer CNN with mini-batch size 1 for vanilla network, LN or BN w/ 4 Things, and BNP. BN and BN Renorm do not train and their results are not plotted. 
    The lines graph the means while the shaded regions graph the ranges of 5 tests.
    }
    \label{fig_bs_1_6}
\end{figure}

We also consider small mini-batch sizes and present the averaged results with mini-batch sizes 6 and 1 for CIFAR10 in Figure \ref{fig_bs_1_6}.
With mini-batch size 6, BNP outperforms BN and the vanilla network.  This demonstrates how small mini-batch sizes can negatively affect BN. We also compare with Batch Renorm and LN in this test.  Batch Renorm and LN perform better than BN and are more comparable to BNP, where BNP converges slightly faster but Batch Renorm achieves the same final accuracy. 
For mini-batch size 1, BN and BN Renorm do not train and are not included in the plot. BNP, LN and the vanilla network all train with mini-batch size 1. BNP and LN converges significantly faster than the vanilla network. They overfit very slightly to have final test accuracy trending just below the vanilla network.   

{\bf Convolutional Neural Networks.} 
We consider a 5-layer CNN as used in \cite{TensorCNN}. 
This network consists of 3 convolution layers, each with a $3 \times 3$ kernel, of filter size $32$-$64$-$32$, followed by two dense layers. 
For this experiment, we also include the version of BN with 4 things/improvements  \citep{Summers2020Four}. These improvements are as follows: for small batch size, BN is combined with GN, while for larger batch-sizes BN is combined with Ghost Normalization; additionally a weight decay regularization on BN's $\gamma, \beta$ terms  and a method to incorporate example statistics during inference are applied \citep{Summers2020Four}. In particular, the combination with GN makes it applicable to mini-batch size 1. 

\begin{figure}[ht]
    \centering
    \begin{subfigure}{.49\textwidth}
    \centering
    \includegraphics[width=1\columnwidth]{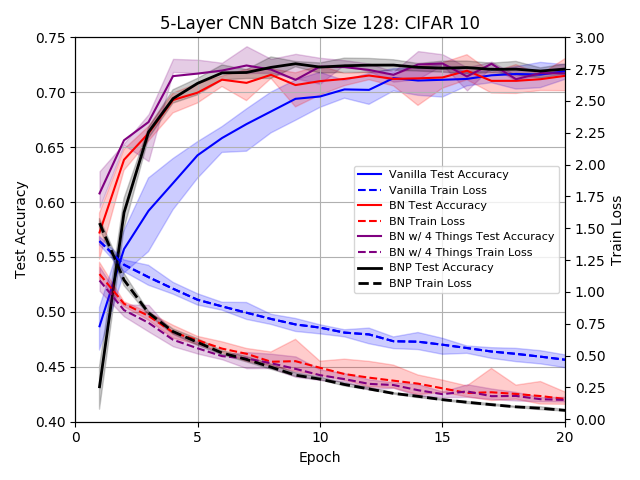}
    \end{subfigure}
    \begin{subfigure}{.49\textwidth}
    \centering
    \includegraphics[width=1\columnwidth]{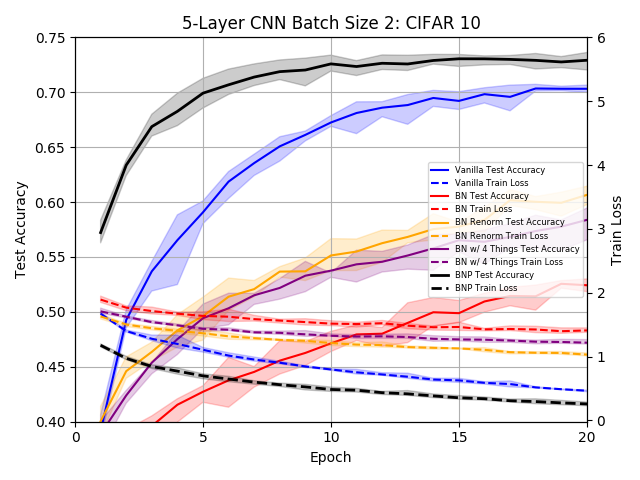}
    \end{subfigure}
    \caption{5-layer CNN for mini-batch size 128 (Left) and mini-batch size 2 (Right): Training loss (dashed lines) and test accuracy (solid lines) for  vanilla network, BN, BN w/ 4 Things, and BNP. For batch-size 2, BN Renorm is also included. The lines graph the means while the shaded regions graph the ranges of 5 tests.
    }
   \label{fig_cnn_128_6}
\end{figure}

We use SGD with the exception of the vanilla and the BN with 4 Things networks where Adam performs better and was used. 
Figure \ref{fig_cnn_128_6} presents the results with 5 tests for CIFAR10 with mini-batch sizes 128 and 2. 
For mini-batch size 128, we find that BN, BN with 4 Things, and BNP all have comparable performance  with BNP converging slightly slower and BN with 4 Things slightly faster than BN at the beginning. They all outperform the vanilla network. 
For mini-batch size 2, BNP significantly outperforms BN, Batch Renorm, and BN with 4 Things. BN underperforms all other methods. However, we note that BN and Batch Renorm work well  for batch sizes as small as 4 and 3 respectively. Compared with the fully connected network, BN appears to be effective in CNNs for much smaller mini-batch size.  

We also test on mini-batch size 1 with 5 test results  given in Figure \ref{fig_bs_1_6} (Right). As expected, BN and BN Renorm do not train and their results are not included in the figure. BNP trains faster and reaches slightly higher accuracy than the vanilla network, while BN with 4 Things significantly underperforms with mini-batch size 1.



Our results demonstrate that BNP works well in the setting of small mini-batch sizes or the online setting. In this situation, it can significantly outperform BN for both fully connected networks and CNNs. It also produces comparable results when larger mini-batch sizes are used.

{\bf Residual Networks.}
We consider 110-layer Residual Networks (ResNet-110) \citep{He15,He16}, and an 18-layer Residual Network (ResNet-18).
  The ResNet-110  has 54 residual blocks, containing two $3 \times 3$ convolution layers each block and was used for CIFAR 10/CIFAR 100 \citep{He15,He16}.
We experiment with both the original ResNet-110 \citep{He15} and its preactivation version \citep{He16}. The ResNet-18 has 8 residual blocks, with  two  $3 \times 3$ convolution layers each block and was used for ImageNet \citep{He15}.
We follow the settings of \citep{He15,He16} by using the data augmentation as in \cite{He15}, the momentum optimizer,  learning rate decay, learning rate warmup, and weight decay; see Appendix B for details.
In particular, we use learning rate decay at epoch  80/120 (CIFAR10), epoch 150/220 (CIFAR100), and epoch 30/60/90 (ImageNet).

\begin{figure}[ht]
    \centering
    \begin{subfigure}{.49\textwidth}
    \centering
    \includegraphics[width=1\columnwidth]{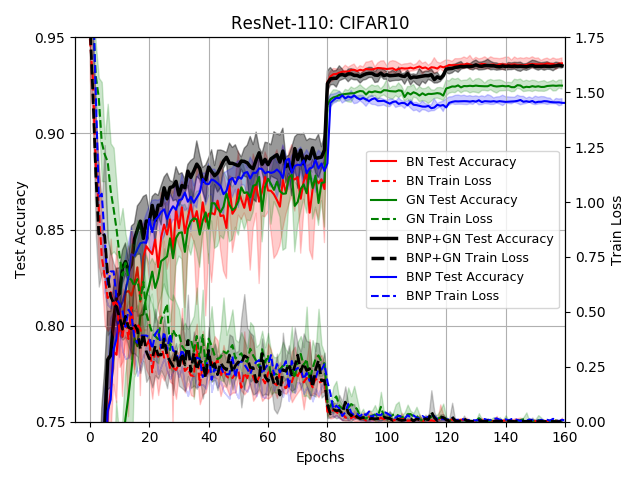}
    \end{subfigure}
    \begin{subfigure}{.49\textwidth}
    \centering
    \includegraphics[width=1\columnwidth]{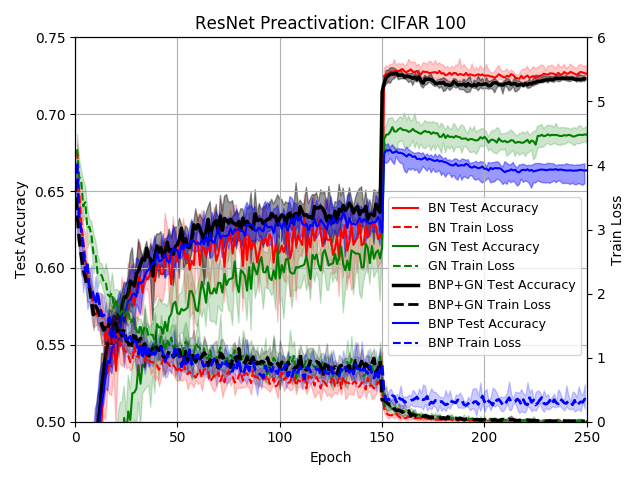}
    \end{subfigure}
    \caption{ResNet-110 for CIFAR10 (Left) and  Preactivation ResNet-110 for CIFAR100 (Right): Training loss (dashed lines) and test accuracy (solid lines) for BN, GN, BNP alone, and BNP+GN.  Both with mini-batch size 128. The lines graph the means while the shaded regions graph the ranges of 5 tests.}
   \label{cifar10_ResNet_20_110_fig}
\end{figure}

One remarkable property of BN when applied to ResNet is its significant increase in test accuracy at learning rate decay. Like most other methods, BNP directly applied to ResNet can not take advantage of the learning rate decay nearly to the same degree as BN, even though BNP converges faster before the learning rate decay; see Figure \ref{cifar10_ResNet_20_110_fig}. We suspect this is due to the lack of scale-invariant property in BNP that is present in BN. Since GN also possess the scale-invariant property,  we may combine BNP with GN; see Section \ref{gn_section}. So, for the ResNet experiment, we apply BNP to ResNet with GN, denoted  by BNP+GN, which is found to benefit much more  significantly from learning rate decay.  For comparison, we also include ResNet with GN in this experiment. 

We test ResNet-110 on CIFAR10   and present the results of 5 runs in  Figure \ref{cifar10_ResNet_20_110_fig} (Left). The test accuracy for BNP+GN  converges faster than BN prior to the first learning rate decay at the 80th epoch and   is comparable to BN after the decay. Both BNP+GN and BN outperform GN significantly.  

We then test preactivation ResNet-110 on CIFAR100 and present the results of 5 runs in  Figure \ref{cifar10_ResNet_20_110_fig} (Right). The test accuracy for BNP+GN also converges faster than BN prior to the first learning rate decay at the 150th epoch but   is slightly lower than BN after the decay. Both BNP+GN and BN  outperform GN significantly.  

\begin{figure}[ht]
    \centering
    \includegraphics[scale=.5]{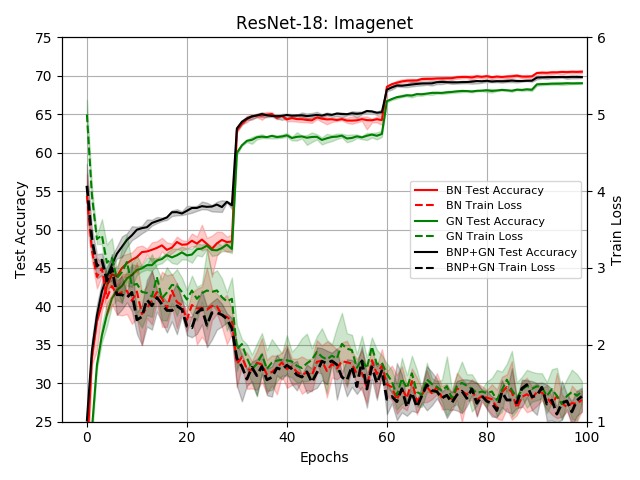}
    \caption{ResNet-18 for ImageNet: Training loss (dashed lines) and test accuracy (solid lines) for BN, GN, and BNP+GN. All have mini-batch size 256.}
   \label{imagenet_resnet18_fig}
\end{figure}

We lastly test ResNet-18  on ImageNet and present the results in  Figure  \ref{imagenet_resnet18_fig}.   The test accuracy for BNP+GN again converges faster than BN prior to the first learning rate decay at the 30th epoch but   is slightly lower than BN after the second decay at the 60th. Again, BNP+GN and BN outperform GN significantly.  

We observe that, for all three ResNets, BNP with the help of GN achieves faster convergence before learning rate decay, while producing comparable final accuracy at the end. With GN's performance lagging, these results can be attributed to BNP and demonstrates the convergence acceleration property of BNP.

\subsection{Exploratory Experiments}
 We present experiments that support our theory that BNP lowers the condition number of the Hessian of the loss function in relation to the condition number of matrix $D$ used in the preconditioning transformation  as detailed in Section 3. We also present a computational timing comparison between BN and BNP.

\begin{figure*}[htbp]
\small
 \begin{tabular}{ccc}
\includegraphics[height=1.9in,width=1.9in]{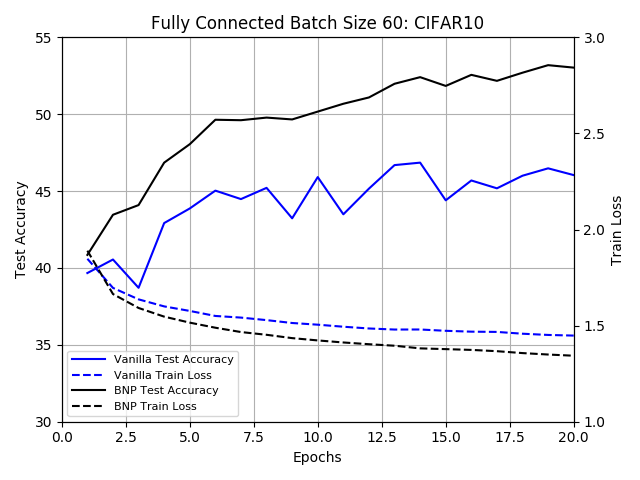} &
\includegraphics[height=1.9in,width=1.9in]{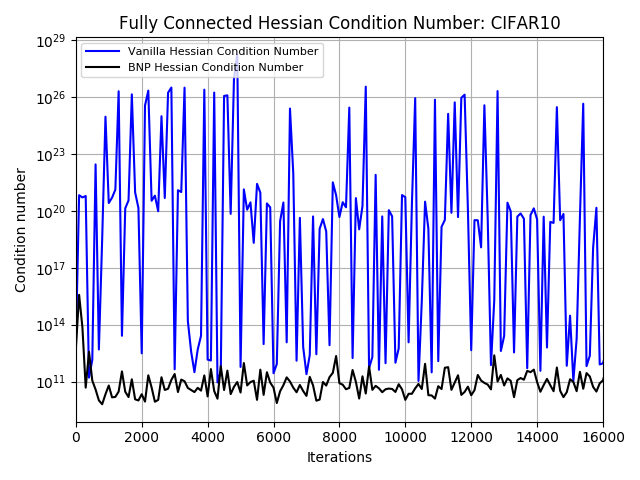} &
\includegraphics[height=1.9in,width=1.9in]{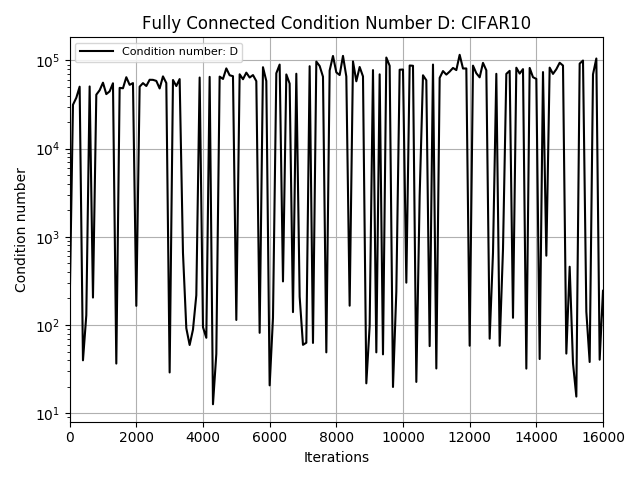}
\end{tabular}

  \caption{We compare the test accuracy (solid lines) and training loss (dashed lines) of a vanilla network with BNP (Left). With this same network we graph and compare the condition number of the Hessian and the preconditioned Hessian by BNP (Center) as well as the condition number of matrix $D$, used in the preconditioning transformation (Right).}
\label{fig_hessian}
\end{figure*}

{\bf Condition Number Analysis:}
We consider a fully connected network consisting of two hidden layers of size 100 and an output layer of size 10. We use the CIFAR10 data set with mini-batch size 60. With this network we compute the condition number of the Hessian with respect to a single weight vector and bias entry at each iteration. In particular, we consider the first weight vector and bias entry in the output layer, that is  $\hat{w}^T=[b_3^{(1)}, w_3^{(1)}] \in \mathbb{R}^{1 \times 101}$, and compute the condition number of the Hessian of the loss function with respect to $\hat{w}$. Results for this condition number and the one for the preconditioned Hessian are given in Figure \ref{fig_hessian} (Center). The preconditioning significantly reduces the condition number of the Hessian. For this same layer, we compute the condition number of matrix $D$. Note that, when $D$ is ill-conditioned, that is when the variances of the activations differ in magnitude, we expect the preconditioning transformation to improve the condition number of the Hessian by approximately $\kappa (D)^2$. For this small network, Figure \ref{fig_hessian} (Right) shows the condition number of $D$ to be in the range $10^2-10^5$. As a result, the BNP transformation reduces the condition number by roughly  $\kappa (D)^2$. We also graph test accuracy and training loss of our Vanilla network versus BNP which confirms accelerated convergence by BNP. These results support our theory.






\begin{figure*}[htbp]
\small
 \begin{tabular}{ccc}
\includegraphics[height=1.9in,width=1.9in]{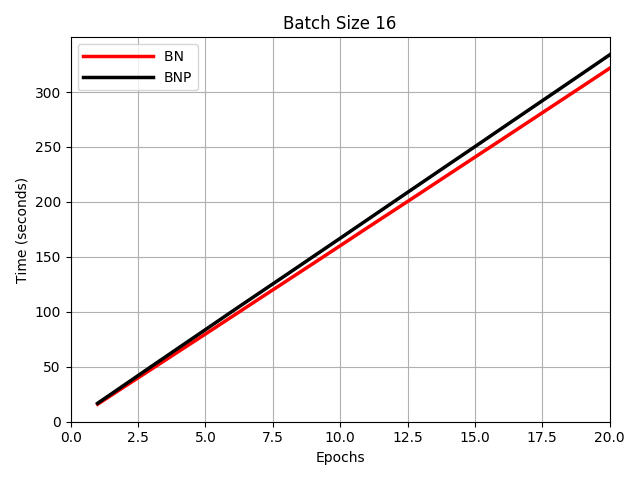} &
\includegraphics[height=1.9in,width=1.9in]{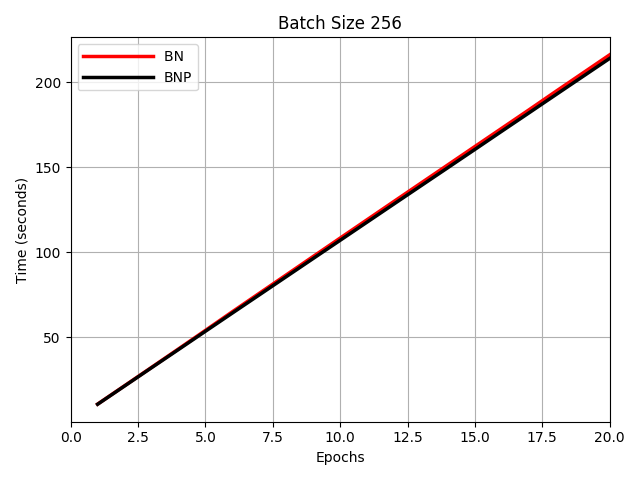} &
\includegraphics[height=1.9in,width=1.9in]{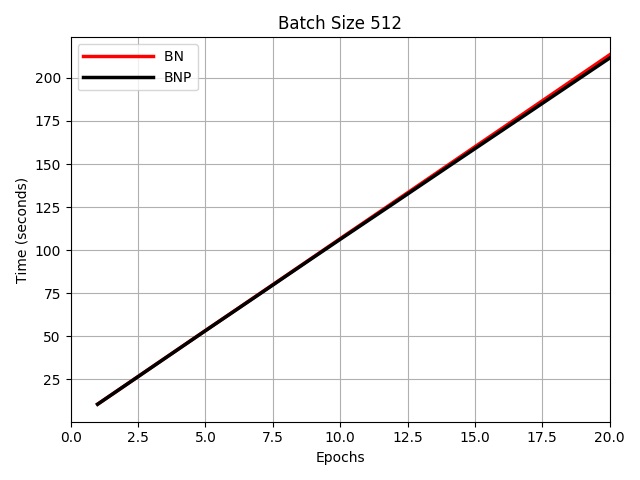}
\end{tabular}

  \caption{Time analysis comparison of BN with BNP. We compare mini-batch sizes of 16, 256, and 512 shown respectively above.}
\label{fig:time_analysis}
\end{figure*}

{\bf Computation Time Analysis:}
We include experiments on the performance time of BN versus BNP over each training epoch. We use the same fully connected network as described in the condition number experiments. These performance time experiments are computed on NVIDIA Tesla V100-SXM2-32GB. We compute the time over each epoch of training and report the cumulative training time over 20 epochs as seen in Figure \ref{fig:time_analysis}. We also report the average time to complete one epoch of training with the respective mini-batch sizes in Table \ref{average_time_table}. We find that for smaller mini-batch size BN is faster than BNP and for larger mini-batch size BNP has a tiny improvement over  BN. Overall, the difference is small and the two methods are quite comparable in computational efficiency. 

\begin{table}[ht]
\begin{center}
\begin{tabular}{ |c|c|c|c| } 
 \hline
 \multicolumn{4}{|c|}{Average Computed Time Over One Epoch}\\
 \hline \hline
  & mini-batch size 16 & mini-batch size 256 & mini-batch size 512 \\ 
 \hline
 BN & 16.08 & 10.81& 10.67 \\ 
 \hline
 BNP & 16.69 &  10.70  &  10.57 \\ 
 \hline
\end{tabular}
\end{center}
\caption{Above we report the average time (in seconds) to complete one training epoch in our small fully connected network given mini-batch sizes 16, 256, and 512.}
\label{average_time_table}
\end{table}

\section{Conclusion}

We have introduced the BNP algorithm that increases the convergence rate of training.  This is done by an implicit change of variables through preconditioned gradient descent. The BNP algorithm is explicitly derived and theoretically supported. It is shown to be equivalent to BN over one training iteration if the back-propagation does not pass through the mini-batch statistics. Its flexibility in using approximate batch statistics allows it to work with small mini-batch sizes.  To summarize the contributions of this work, our theory provides an insight on why BN works and how it should be applied to CNNs, and our algorithm provides a practical alternative to BN when a very small mini-batch size is used. 

 \medskip






\acks{This research was supported in part by NSF under the grants  DMS-1620082 and  DMS-1821144. We would like to thank three anonymous referees for many constructive comments and suggestions that have significantly improved the paper.  We would also like to thank the University of Kentucky Center for Computational Sciences and Information Technology Services Research Computing for their support and use of the Lipscomb Compute Cluster and associated research computing resources.}

\appendix

\section*{Appendix A. Proof of Theorems}
In this appendix, we present the proofs of theorems not proven in the discussions of the main text. In particular, Theorem \ref{precond_hess}, Proposition \ref{flops}, and Lemma \ref{CNN_notation_theorem} are omitted for this reason.
For all remaining proofs, we restate the original theorems for completeness. \\

\noindent \textbf{Theorem \ref{thm:hessian_convergence}}
\textit{
Consider a loss function $\mathcal{L}$ with continuous third order derivatives and with a positively scale-invariant property. If $\theta=\theta^*(t)$ is a positively scale-invariant manifold at local minimizer $\theta^*$, then the null space of the Hessian $\nabla^2 \mathcal{L} (\theta^* (t))$   contains at least the column space ${\rm Col}\left( D_t \theta^*(t)\right)$. Furthermore, for the gradient descent iteration $\theta_{k+1}=\theta_{k}-\alpha \nabla \mathcal{L} (\theta_k)$, let $\theta^*_k$ be the local minimizer closest to $\theta_k$, i.e. $\theta^*_k=\theta^* (t_k)$ where $t_k=\text{argmin}_{t>0} \|\theta_k - \theta^* (t)\|$ and assume that  the null space of $\nabla^2 \mathcal{L} (\theta^* (t))$ is equal to ${\rm Col}\left( D_t \theta^*(t)\right)$. Then, for any $\epsilon >0$, if our initial approximation $\theta_0$ is such that $||\theta_0 - \theta^*_0||$ is sufficiently small, then  we have, for all $k \geq 0$,
$$||\theta_{k+1}-\theta^*_{k+1} ||\leq (r+\epsilon)||\theta_k-\theta^*_k||,$$ where
$r=\max\{|1-\alpha \lambda^{*}_{min}|, ||1-\alpha \lambda_{max}||\}$ 
and $\lambda^*_{min}$ and $\lambda_{max}$ are the smallest nonzero eigenvalue and the largest eigenvalue of $\nabla^2 \mathcal{L} (\theta^*_k)$, respectively.
}
\begin{proof} The proof of the first part is contained in \S 3.2. Here we prove the second part. 

Since $\theta^*_k=\theta^*(t_k)$ is the local minimizer closest to $\theta_k$, i.e. $\|\theta_k-\theta^*(t_k)\| = \min_t  \|\theta_k-\theta^*(t)\|$, taking derivation in $t$ results in  $(\theta_k-\theta^*(t_k))^T D_t \theta^*(t_k)=0$. Then $(\theta_k-\theta^*_k)$ is perpendicular to the null space of the Hessian. 
Furthermore, $||\theta_{k+1}-\theta^*_{k+1}||  \leq ||\theta_{k+1}-\theta^*_{k}||$.

Using  $\theta_{k+1}=\theta_{k}- \alpha \nabla \mathcal{L}(\theta_k)$, we have
\begin{align}
    ||\theta_{k+1}-\theta^*_{k+1}||  &\leq ||\theta_{k+1}-\theta^*_{k}|| \notag \\ 
    &= ||\theta_{k}- \alpha \nabla \mathcal{L}(\theta_k) -\theta^*_{k}|| \notag \\
   &= ||\theta_{k}-\theta^*_{k}- \alpha (\nabla \mathcal{L}(\theta_k)-\nabla\mathcal{L}(\theta^*_k)) || \label{taylor_approx}
\end{align}
Since $\mathcal{L}$ has continuous third order derivatives, we have $\nabla \mathcal{L} (\theta_k)=\nabla \mathcal{L} (\theta^*_k) +\nabla^2 \mathcal{L} (\theta^*_k) (\theta_k-\theta^*_k) +e_k$, where $e_k$ is the remainder with  $||e_k|| \leq C ||\theta_k-\theta^*_k||^2$ for some $C$. Substituting into Equation (\ref{taylor_approx}), we get
\begin{align}
    ||\theta_{k+1}-\theta^*_{k+1}||& \leq ||\theta_{k}-\theta^*_{k}- \alpha \nabla^2 \mathcal{L}(\theta^*_k)(\theta_{k}-\theta^*_{k})||+\alpha ||e_k|| \notag  \\
    &= ||(I- \alpha \nabla^2 \mathcal{L}(\theta^*_k) )(\theta_{k}-\theta^*_{k})||+\alpha||e_k|| \\ 
    & \leq \max_{\lambda_i\neq 0 }|1-\alpha \lambda_i|\cdot ||\theta_k - \theta^*_k||+\alpha C ||\theta_k-\theta^*_k||^2, \notag
\end{align}
where $\lambda_1, ... ,\lambda_n$ are eigenvalues of $\nabla^2 \mathcal{L}(\theta^*_k)$ and we have used that $\theta_k-\theta^*_k$ is perpendicular to the null space in the last inequality. 
Notice $$\max_{\lambda_i\neq 0 }|1-\alpha \lambda_i| = \max \{|1-\alpha \lambda^*_{\min}|,|1-\alpha \lambda_{\max}| \} =r.$$ We have 
\begin{equation}
||\theta_{k+1}-\theta^*_{k+1}||  \leq r \cdot ||\theta_{k}-\theta^*_{k}||+\alpha C ||\theta_k-\theta^*_k||^2=\big(r+\alpha C||\theta_{k}-\theta^*_{k}||\big)\cdot ||\theta_{k}-\theta^*_{k}||.
\label{eq:r_c_bound}
\end{equation}
Clearly, $r<1$. For any $\epsilon>0$ with $r+\epsilon \leq1$, let $\delta=\frac{ \epsilon}{\alpha C}$. Then, if $||\theta_{k}-\theta^*_{k}||<\delta$, we have 
\[
||\theta_{k+1}-\theta^*_{k+1}||  \leq\big(r+\alpha C\delta \big)\cdot ||\theta_{k}-\theta^*_{k}|| 
 \leq  ||\theta_{k}-\theta^*_{k}|| < \delta.
\]
Thus, if $ ||\theta_0-\theta^*_0|| < \delta,$ by an induction argument, we have  $||\theta_{k}-\theta^*_{k}||<\delta$ for all $k$. 
Hence, 
$$||\theta_{k+1}-\theta^*_{k+1}|| \leq (r+\epsilon)||\theta_{k}-\theta^*_{k}||.$$
\end{proof}

\textbf{Proposition \ref{grad_hessian_theorem}.}
\textit{Consider a loss function $L$  defined from the output of a fully connected multi-layer neural network for a single network input $x$. Consider the weight and bias parameters $w_{i}^{(\ell)}, b_{i}^{(\ell)}$ at the $\ell$-layer and 
write $L = L\left(a_{i}^{(\ell)} \right) = L\left( \widehat{w}^{T}\widehat{h} \right)$ as a function of the parameter $\widehat{w}$ through $a_{i}^{(\ell)}$ as in (\ref{eq:ai}).  
When training over a mini-batch of $N$ inputs,
let $\{h_{1}^{(\ell-1)}, h_{2}^{(\ell-1)}, \hdots, h_{N}^{(\ell-1)} \}$  be the associated $h^{(\ell-1)}$
and let $\widehat{h}_j=\begin{bmatrix}
       1  \\
        h_j^{(\ell-1)}
      \end{bmatrix}
 \in \mathbb{R}^{(n_{\ell}+1)\times 1}$. Let $$\mathcal{L} =\mathcal{L} (\widehat{w}) := \frac{1}{N}\sum_{j = 1}^{N}L\left(\widehat{w}^T\widehat{h}_j \right)$$ be the mean loss over the mini-batch. Then, its Hessian with respect to $\hat{w}$  is 
\begin{equation*}
    \nabla^2_{\widehat{w}} \mathcal{L} (\widehat{w}) =  \widehat{H}^{T}S\widehat{H}  
\end{equation*}
    \text{where } 
\begin{equation*} \widehat{H}=[e, \; H], 
     {H} = \begin{bmatrix}
        h_{1}^{(\ell-1)^T} \\
       \vdots\\
        h_{N}^{(\ell-1)^T}
      \end{bmatrix} 
      \text{ and } S = \frac{1}{N} \text{diag}\left(L''\left(\widehat{w}^T\widehat{h}_j\right)\right).
\end{equation*}
with all off-diagonal elements of $S$ equal to 0.}

\begin{proof}
First, we have $\nabla_{\widehat{w}} L\left( \widehat{w}^{T}\widehat{h} \right)=L' \left( \widehat{w}^{T}\widehat{h} \right) \widehat{h}$. Then, taking the gradient of $\mathcal{L}$ with respect to $\widehat{w}$, 
\begin{align}
    \nabla_{\widehat{w}}\mathcal{L} = \frac{1}{N}\sum_{j=1}^{N}{L}'\left(\widehat{w}^{T}\widehat{h}_j\right)\widehat{h_j}.
    \label{gradient_eqn2}
\end{align}
Furthermore, 
$\nabla^2_{\widehat{w}} L\left( \widehat{w}^{T}\widehat{h} \right)=L'' \left( \widehat{w}^{T}\widehat{h} \right) \widehat{h} \widehat{h}^T$.
Applying this to  (\ref{gradient_eqn2}), we obtain the desired Hessian
\begin{align*}
    \nabla^{2}_{\widehat{w}}\mathcal{L} =& \frac{1}{N}\sum_{j=1}^{N} {L}''\left(\widehat{w}^T\widehat{h}_j \right)\widehat{h}_{j}\widehat{h}_{j}^{T} 
    \nonumber \\
    =& \widehat{H}^{T}S\widehat{H}
    \label{hessian_eqn2}
\end{align*}
where we note that $\widehat{H}^{T}=[\widehat{h}_{1}, \widehat{h}_{2}, \cdots, \widehat{h}_{N}]$.
\end{proof}

Before proving Theorem \ref{thm:pcond}, we first present a result of  \cite{van1969condition} as a lemma. 

\textbf{Lemma A1.} \citep{van1969condition} \textit{
 Let $A\in \mathbb{R}^{m\times n }$ have  full column rank and let $D={\rm diag}\{\|a_1\|^{-1}, \cdots, \|a_n\|^{-1}\}$, where $a_i$ is the $i$-th column of $A$. We have 
\begin{equation*}
 \kappa(AD)  \le \sqrt{n} \min_{D_0 \mbox{ is diagonal}} \kappa (A D_0).
 \label{sluis_ori}
 \end{equation*}
}

\textbf{Theorem \ref{thm:pcond}.} 
\textit{Let $\widehat{H}=[e, H]$ be the extended hidden variable matrix,  $\widehat{G}$ the normalized hidden variable matrix, $U$ the centering transformation matrix, and $D$ the variance normalizing matrix defined by (\ref{hessian_eqn1}), (\ref{PD_eqn}), and (\ref{Gmatrix}). Assume $\widehat{H}$ has full column rank. We have}
 \begin{equation*}
 \kappa(\widehat{H}U) \le   \kappa(\widehat{H}). \end{equation*}
\textit{This inequality is strict if $\mu_H \ne 0$ and is not orthogonal to $x_{\max}$, where $x_{\max}$ is an eigenvector corresponding to the largest eigenvalue of the sample covariance matrix $\frac{1}{N-1}(H-e\mu_{H}^T)^T(H-e\mu_{H}^T)$ (i.e. principal component). Moreover,}
\begin{equation*}
 \kappa(\widehat{G}) =\kappa (\widehat{H} U D) \le \sqrt{n_{\ell-1}+1} \min_{D_0 \mbox{ is diagonal}} \kappa (\widehat{H} U D_0).
 \end{equation*}

\begin{proof}
We show  $$\kappa(\widehat{H}U) =\sqrt{\frac{\lambda_{max}(U^T\widehat{H}^T \widehat{H}U)}{\lambda_{min}(U^T\widehat{H}^T \widehat{H} U)}}\leq \sqrt{\frac{\lambda_{max}(\widehat{H}^T \widehat{H})}{\lambda_{min}(\widehat{H}^T \widehat{H})}}=\kappa(\widehat{H}),$$
by proving $\lambda_{max}(U^T\widehat{H}^T \widehat{H}U) \leq \lambda_{max}(\widehat{H}^T \widehat{H})$ and $\lambda_{min}(U^T\widehat{H}^T \widehat{H}U) \geq \lambda_{min}(\widehat{H}^T \widehat{H})$, where $\lambda_{min}(A)$ and $\lambda_{max}(A)$ denote, respectively, the minimum and maximum eigenvalues of a matrix $A$. We use the Courant-Fisher minimax characterization of eigenvalues, i.e. 
\begin{align*}
\lambda_{max}(U^T\widehat{H}^T \widehat{H}U)&=\text{max}_{z \neq 0}\frac{ 
     z^T U^T\widehat{H}^T \widehat{H}Uz}{z^T z}
     \end{align*}
      and \begin{align*}
\lambda_{min}(U^T\widehat{H}^T \widehat{H}U)&=\text{min}_{z \neq 0}\frac{     z^T U^T\widehat{H}^T \widehat{H}Uz}{z^T z}
     \end{align*}
     
Write  
\[
U^T\widehat{H}^T \widehat{H}U=\begin{bmatrix}
     N & 0 \\
     0 & (H^T-\mu_He^T)(H-e\mu_H^T)
     \end{bmatrix}
\]
as a $2 \times 2$ block matrix, with $H \in \mathbb{R}^{N \times m}.$
Write $z=[t,\; x^T]^T$,    where $ t \in \mathbb{R}$ and $x \in \mathbb{R}^m$ and we simplify to obtain
     \begin{align}
       z^T U^T\widehat{H}^T \widehat{H}Uz &= \begin{bmatrix}
     t & x^T
     \end{bmatrix}U^T\widehat{H}^T \widehat{H}U\begin{bmatrix}
     t\notag  \\
     x 
     \end{bmatrix} \\
     &=t^2N+x^T(H^T-\mu_He^T)(H-e\mu_H^T)x\notag \\
     &=t^2N+x^TH^THx-x^TH^Te\mu_H^Tx-x^T\mu_He^THx+x^T\mu_He^T e\mu_H^Tx\notag \\
     &=t^2N+x^TH^THx-N(x^T\mu_H)^2 \notag \\ 
    &=N(t^2-(x^T\mu_H)^2)+x^TH^THx, \label{HUtransposeRay}
    \end{align}
    where we have used $H^Te=N \mu_H$ and $e^Te=N$.
    Similarly, 
we have 
 \begin{align}
        z^T\widehat{H}^T \widehat{H}z &= \begin{bmatrix}
     t & x^T
     \end{bmatrix}\widehat{H}^T \widehat{H}\begin{bmatrix}
     t \notag \\
     x 
     \end{bmatrix}\\
     &=t^2N+te^THx+x^TH^Tet+x^TH^THx \notag\\
     &=t^2N+2tNx^T\mu_H+x^TH^THx\notag\\
     &=N(t^2+2tx^T\mu_H)+x^TH^THx \label{Htranspose_maxray}\\
     &= N(t+x^T\mu_H)^2-N(x^T\mu_H)^2+x^TH^THx. \notag
    \end{align}
    %

We now prove $\lambda_{max}(U^T\widehat{H}^T \widehat{H}U) \leq \lambda_{max}(\widehat{H}^T \widehat{H})$ with a strict inequality if $\mu_H^T x_{\max} \ne 0$.     
The maximum eigenvalue of $U^T\widehat{H}^T \widehat{H}U$ is either $N$ or  the maximum eigenvalue of $(H^T-\mu_He^T)(H-e\mu_H^T)$. In the first case,  using equation (\ref{Htranspose_maxray}), we have  
  \begin{align} \lambda_{max}(U^T\widehat{H}^T\widehat{H}U) 
  &=N \notag \\
    &\le \frac{N(1 +2 \mu_H^T\mu_H)+\mu_H^TH^TH\mu_H}{1+\mu_H^T\mu_H} \label{ine} \\
      & \le \max_{[t, x^T] \neq 0}\frac{N(t^2+2tx^T\mu_H)+x^TH^THx}{t^2+x^Tx}\notag \\ 
    &= \lambda_{max}(\widehat{H}^T\widehat{H}),\notag 
    \end{align}
where the inequality (\ref{ine}) is strict if $\mu_H \ne 0$. 
In the second case, using (\ref{HUtransposeRay}) with $t=0$, we have 
  \begin{align} \lambda_{max}(U^T\widehat{H}^T\widehat{H}U) 
  &= \frac{x_{\max}^T (H^T-\mu_He^T)(H-e\mu_H^T) x_{\max}}{x_{\max}^T x_{\max}} \notag \\
  &= \frac{-N (x_{\max}^T \mu_H)^2 + x_{\max}^T H^TH x_{\max}}{x_{\max}^T x_{\max}} \notag \\
 &\le \frac{N (x_{\max}^T \mu_H)^2-N (x_{\max}^T \mu_H)^2 + x_{\max}^T H^TH x_{\max}}{x_{\max}^T x_{\max}} \label{ine2} \\
      & \le \max_{[t, x^T] \neq 0}\frac{ N(t+x^T\mu_H)^2-N(x^T\mu_H)^2+x^TH^THx}{t^2+x^Tx} \notag \\ 
    &= \lambda_{max}(\widehat{H}^T\widehat{H}), \notag
    \end{align}
where  the inequality (\ref{ine2}) is strict if $x_{\max}^T \mu_H \ne 0$.

Next we show $\lambda_{min}(U^T\widehat{H}^T\widehat{H}U) \geq \lambda_{min}(\widehat{H}^T\widehat{H})$. First, using  (\ref{Htranspose_maxray}), we have 
\begin{align}\lambda_{min}(U^T\widehat{H}^T\widehat{H}U)&=\min_{[t, x^T] \neq 0}\frac{N(t^2-(x^T\mu_H)^2)+x^TH^THx}{t^2+x^Tx}\notag \\
    &=\min_{t\ge 0, \; x^T \mu_H \le 0,\; [t, x^T] \neq 0}\frac{N(t^2-(x^T\mu_H)^2)+x^TH^THx}{t^2+x^Tx}   \label{lammin}
    \end{align}
    where (\ref{lammin}) holds since the function is not dependent on the sign of $t$ or $x^T\mu_H$. 
Let $t, x$ be such that $t\ge 0$ and $x^T \mu_H \le 0$. We discuss two cases. First consider the case $0 \geq x^T\mu_H \geq -2 t$. This implies $|t+x^T\mu_H| \leq t$, and 
\[
    \frac{Nt^2-N(x^T\mu_H)^2+x^TH^THx}{t^2+x^Tx} \ge \frac{N(t+x^T\mu_H)^2-N(x^T\mu_H)^2+x^TH^THx}{t^2+x^Tx}.
\]
Now, consider the case $x^T\mu_H \le - 2t\le 0$. Let $s=-t -x^T\mu_H$. Then $s\ge t\ge 0$ and $s^2 \ge t^2$.  This implies 
\begin{align*}
         \frac{Nt^2-N(x^T\mu_H)^2+x^TH^THx}{t^2+x^Tx} &= \frac{N(s+x^T\mu_H)^2-N(x^T\mu_H)^2+x^TH^THx}{t^2+x^Tx} \\
        &\ge  \frac{N(s+x^T\mu_H)^2-N(x^T\mu_H)^2+x^TH^THx}{s^2+x^Tx}.   
\end{align*}
In either case, 
\begin{align*}
        \frac{Nt^2-N(x^T\mu_H)^2+x^TH^THx}{t^2+x^Tx} 
        &\ge \min_{[s,x^T] \neq 0} \frac{N(s+x^T\mu_H)^2-N(x^T\mu_H)^2+x^TH^THx}{s^2+x^Tx} \\
        &= \lambda_{min}(\widehat{H}^T\widehat{H}).
\end{align*}

Hence, it follows from this and (\ref{lammin}) that   $\lambda_{min}(U^T\widehat{H}^T\widehat{H}U) \ge \lambda_{min}(\widehat{H}^T\widehat{H}).$
This gives us the desired result that $\kappa(\widehat{H}U) \leq \kappa(\widehat{H}).$
For the second bound, we note that each column of $\widehat{H} U D$ has norm $\sqrt{N}$ so each column of $\widehat{H} U (D/\sqrt{N})$ has norm $1$. By Lemma A1,  \[
\kappa(\widehat{H} U (D/\sqrt{N})) \le \sqrt{n_{\ell-1}+1} \min_{D_0 \mbox{ is diagonal}} \kappa (\widehat{H} U D_0).
\]
Since $\kappa(\widehat{H} U (D/\sqrt{N}))=\kappa(\widehat{H} U D)$, the bound is proved.
\end{proof}

Next, we show Proposition \ref{randommatrix}. We need to use Corollary 2.2 of \cite{seginer_2000}, which is stated here. 

\textbf{Lemma A2.} \citep{seginer_2000} \textit{There exists a constant $C$ such that, for any $m,n$, $k \le 2 \log \max\{m,n\}$, and any $G=[g_{ij}] \in \R^{m\times n}$ where $g_{ij}$ are iid zero mean random variables, the following inequality holds:
\[
E\left[ \|G\|^k\right] 
\le C^k 
\left(E\left[\max_{1\le i \le m} \|g_i\|^k\right]+ 
E\left[\max_{1\le j \le n} \|h_j\|^k\right]\right)
\]
where $g_i$ is the $i$th row of $A$ and $h_j$ is the $j$th column of $A.$
}

\textbf{Proposition \ref{randommatrix}.} \textit{Let $n_{\ell-1}\ge 3$ and assume the entries of the normalized hidden variable matrix $G\in \R^{N\times n_{\ell-1}}$ are iid  random variables with zero mean and unit variance. Then the expectation of the norm of  $(1/q) \widehat{G}= (1/q)\left[e,\; G\right]$ is bounded by $C \sqrt{N} $ for some constant $C$ independent of $n_{\ell-1}$, where $q^2 = {\max\{{n_{\ell-1}}/{N}, 1\}}$. 
}

\begin{proof}
Applying Lemma A2 with $k=2$ to $G$ here, we have 
$E\left[ \|g_i\|^2\right] =n_{\ell-1}$ and $E\left[ \|h_j\|^2\right] =N$ for $1\le i \le N$ and $1\le j \le n_{\ell-1}$. Since all $g_i$  for $1\le i \le N$ have the identical distribution, $E\left[\max_{1\le i \le N} \|g_i\|^2\right]= n_{\ell-1}$. Similarly, $E\left[\max_{1\le j \le m} \|h_j\|^2\right]=N$. Therefore,  
 \[
 E\left[ \|G\|^2\right] 
\le C_1^2 
\left(N+n_{\ell-1}\right) \le 2C_1^2 \max\{N,n_{\ell-1}\}.
\]
Thus 
\[
(1/q)E\left[ \|\widehat{G} \|\right]=  (1/q) \max\{\sqrt{N}, E\left[\|G\|\right]\}
\le  (1/q) \sqrt{2}C_1 \max\{\sqrt{N}, \sqrt{n_{\ell-1}}\} =\sqrt{2}C_1 \sqrt{N}.
 \]
 The theorem follows with $C= \sqrt{2}C_1.$
\end{proof}

\textbf{Proposition \ref{bnp_equiv_theorem}.} \textit{A post-activation BN network defined in (\ref{post_eqn}) with $\mathcal{B}_{0,1}\left( \cdot \right)$ is equivalent to  a vanilla network (\ref{vanilla_eqn})  with $\widehat{\theta} = \{\widehat{W}, \widehat{b}\}$ where $\widehat{W} = W^{(\ell)}\text{diag}\left(\frac{1}{\sigma_H}\right)$ and $\widehat{b} = b^{(\ell)} - W^{(\ell)}\text{diag}\left(\frac{\mu_H}{\sigma_H}\right) $.  Furthermore, one step of training of BN with $\mathcal{B}_{0,1}\left(\cdot \right)$ without passing the gradient through $\mu_H$, $\sigma_H$ is equivalent to one step of BNP training of (\ref{vanilla_eqn}) with $\widehat{W}^T, \widehat{b}$.
}

\begin{proof}
    Examining (\ref{post_eqn}) with $\beta=0$ and $\gamma = 1$, we have 
    \begin{align}
        h^{(\ell)} =& g\left(W^{(\ell)}\mathcal{B}_{0,1}\left(h^{(\ell-1)}\right) + b^{(\ell)}\right) \label{BN01}\\
        =& g\left(W^{(\ell)}\text{diag}\left(\frac{1}{\sigma_H}\right)\left(h^{(\ell-1)} - \mu_H \right) + b^{(\ell)}\right) \nonumber \\
        =& g\left( \widehat{W} h^{(\ell-1)} + \widehat{b}\right). \label{vanilaBN}
    \end{align}
    This proves the first part. Note that training of the BN network is done through gradient descent in  $W^{(\ell)}, b^{(\ell)}$ of (\ref{BN01}).  Furthermore, 
For $U, D$ and $P$  defined  in (\ref{PD_eqn}), using $\widehat{b}^T = b^{(\ell)^T} - \text{diag}\left(\frac{\mu_H}{\sigma_H}\right) W^{(\ell)^T} $, we have 
    \begin{equation*}
        \begin{bmatrix}
            \widehat{b}^{T}\\
            \widehat{W}^T
        \end{bmatrix} =
        U\begin{bmatrix}
           b^{(\ell)^T}\\
           \widehat{W}^T
        \end{bmatrix} =
        UD\begin{bmatrix}
           b^{(\ell)^T}\\
           W^{(\ell)^T}
        \end{bmatrix}=
        P\begin{bmatrix}
           b^{(\ell)^T}\\
           W^{(\ell)^T}
        \end{bmatrix}
    \end{equation*}
    which is the BNP preconditioning transform (\ref{PD_eqn}) applied to each column of the  matrix $\begin{bmatrix}
            \widehat{b}^{T}\\
            \widehat{W}^T
        \end{bmatrix}$.  Thus, one step of BNP gradient descent in $\widehat{\theta} = \{\widehat{W}, \widehat{b}\}$ of (\ref{vanilla_eqn}) is equivalent to one step of gradient descent in   $W^{(\ell)}, b^{(\ell)}$, which is a gradient descent step for the BN network (\ref{BN01}).
\end{proof}
\noindent \textbf{Proposition \ref{CNN_hessian_theorem}} 
\textit{Consider a CNN loss function $L$ for a single input tensor and write   ${L} ={L}(\mathbf{a}( \cdot, \cdot, d))$  as a function of the convolution kernel $\mathbf{w}( \cdot, \cdot, \cdot, d)$ where $$\mathbf{a}( \cdot, \cdot, d){=\sum_{c=1}^{c_l} \text{Conv} (\mathbf{h}^{(l)}\left(\cdot , \cdot, c),  w(\cdot, \cdot, c,d)\right)}+b_d.$$ When training over a mini-batch with $N$ inputs, let the associated hidden variables 
$\mathbf{h}$ of layer $\ell$ be  $\{ \mathbf{h}^{(1)}, \mathbf{h}^{(2)}, ...,\mathbf{h}^{(N)}\}$ 
and let the associated output of layer $\ell$ be  $\{ \mathbf{a}^{(1)}, \mathbf{a}^{(2)}, ...,\mathbf{a}^{(N)}\}$.
Let $\mathcal{L} =\frac{1}{N}\sum_{j=1}^N L(\mathbf{a}^{(j)}( \cdot, \cdot, d))$ be the mean loss over the mini-batch. Then
\begin{align*}
  \nabla_{\widehat{w}}^2 \mathcal{L}=  \mathcal{\widehat{H}}^TS\mathcal{\widehat{H}},
\end{align*}}
where
\[
S=\frac{1}{N} \left[\begin{array}{cccc}
     \frac{\partial^2 L}{\partial v_a^2}(v_a^{(1)})&&& \\ 
     &\frac{\partial^2 L}{\partial v_a^2}(v_a^{(2)})&&\\
    &&\ddots & \\ &&& \frac{\partial^2 L}{\partial v_a^2}(v_a^{(N)})\\
\end{array}\right], \]
       $v_a:=\text{vec}(\mathbf{a}( \cdot, \cdot, d)) 
       \;\mbox{ and }\; v_a^{(j)}:=\text{vec}\left(\mathbf{a}^{(j)}( \cdot, \cdot, d)\right).$



\begin{proof} For ease of notation, we rewrite the function  ${L} ={L}(\mathbf{a}( \cdot, \cdot, d))$  as  ${L} ={L}(v_a)$. 
First, for a single element $v_a^{(j)}:= \text{vec}(\mathbf{a}^{(j)}(\cdot, \cdot, d))$ in the mini-batch, we have 
$\displaystyle{\frac{\partial L}{\partial \widehat{w}}(v_a^{(j)})=\frac{\partial L}{\partial v_a}(v_a^{(j)})\frac{\partial v_a^{(j)}}{\partial \widehat{w}}}$.
Further, considering $\mathcal{L}$ with respect to $\widehat{w},$
\begin{align*}
  \frac{\partial \mathcal{L}}{\partial \widehat{w}}&= \frac{1}{N}\sum_{j=1}^N  \frac{\partial L}{\partial  \widehat{w}}\\
   &= \frac{1}{N}\sum_{j=1}^N  \frac{\partial L}{\partial v_a} (v_a^{(j)}) \frac{\partial v_a^{(j)}}{\partial \widehat{w}}.
 \end{align*}
 

 
Further, $\displaystyle{\nabla^2_{\widehat{w}} L=\left(\frac{\partial v_a^{(j)}}{\partial \widehat{w}}\right)^T  \frac{\partial^2 L}{\partial v_a^2}(v_a^{(j)}) \frac{\partial v_a^{(j)}}{\partial \widehat{w}}}$, which gives
 \begin{align*}
      \nabla^2_{\widehat{w}} \mathcal{L}&=\frac{1}{N} \sum_{j=1}^N
      \left(\frac{\partial v_a^{(j)}}{\partial \widehat{w}}\right)^T  \frac{\partial^2 L}{\partial v_a^2}(v_a^{(j)}) \frac{\partial v_a^{(j)}}{\partial \widehat{w}}\\
      &=  \left[\begin{array}{cccc}
       \frac{\partial v_a^{(1)}}{\partial \widehat{w}}^T & \frac{\partial v_a^{(2)}}{\partial \widehat{w}}^T & \hdots & \frac{\partial v_a^{(N)}}{\partial \widehat{w}} ^T 
 \end{array}\right]      
S
\left[\begin{array}{cccc}
      \frac{\partial v_a^{(1)}}{\partial \widehat{w}} \\ \frac{\partial v_a^{(2)}}{\partial \widehat{w}} \\ \vdots \\ \frac{\partial v_a^{(N)}}{\partial \widehat{w}}  
\end{array}\right] \\
&= \left(\frac{\partial \widehat{a}}{\partial \widehat{w}}\right)^T S \,  \, \frac{\partial \widehat{a}}{\partial \widehat{w}} \\
&= \mathcal{\widehat{H}}^T S \mathcal{\widehat{H}},
\end{align*}
where $\frac{\partial \widehat{a}}{\partial \widehat{w}}=\mathcal{\widehat{H}}$.
\end{proof} 
\noindent \textbf{Proposition \ref{CNN_error_bound_proposition}} 
\textit{Under the notation defined in Proposition \ref{CNN_hessian_theorem}, the error $|\mu(t,a,p)-\mu (p)|$ of approximating $\mu(t,q,p)$, the mean calculated for the BNP transformation, as defined in (\ref{eq:CNN_mean_submatrix}), by $\mu (p)$, as in (\ref{eq:CNN_approx_mean}), the mean over the entire feature map, is bounded by 
$$\displaystyle{(k-1)\left(\frac{1}{2r}+ \frac{1}{2s} -\frac{k-1}{4rs}\right)\max_{u,v,i}{|h^{(i)}(u,v,p)|}}.$$ }
\begin{proof}
Using the mean computed over the entire mini-batch as in (\ref{eq:CNN_approx_mean}) and the BNP mean as in (\ref{eq:CNN_mean_submatrix}), we get an exact error of
\begin{equation} \label{exact_error_CNN_mean}
|\mu(t,a,p)-\mu (p)| =\frac{1}{Nrs}\Bigg|\sum_{i=1}^{N}\sum_{u=t-\frac{k-1}{2}}^{r+t-\frac{k-1}{2}-1}\sum_{v=a-\frac{k-1}{2}}^{s+a-\frac{k-1}{2}-1} h^{(i)}(u,v,p)-\sum_{i=1}^{N}\sum_{u=1}^{r}\sum_{v=1}^{s} h^{(i)}(u,v,p)\Bigg|,\end{equation}
where $h$ is 0 for any indices outside of its bounds, as $h$ is zero padded. Note this subtraction of a subtensor of $h$ of size $r \times s \times p$ from padded $h$ of size $(r+\frac{k-1}{2}) \times (s+\frac{k-1}{2}) \times p$ gives a difference of at most $\frac{k-1}{2}$ columns and $\frac{k-1}{2}$ rows of the feature map. 
Thus an upper bound for this error is given by summing over the $\frac{k-1}{2}(r+s)N$ elements in the $\frac{k-1}{2}$ boundary columns and rows of each feature map, and subtracting all elements double counted, of which there are $N(\frac{k-1}{2})^2$. Since $|h^{(i)}(u,v,p)| \leq \max_{u,v,i}{|h^{(i)}(u,v,p)|}$, we bound (\ref{exact_error_CNN_mean}) by $$\displaystyle{\frac{(k-1)(r+s)-\frac{(k-1)^2}{2}}{2rs}\max_{u,v,i}{|h^{(i)}(u,v,p)|}=\displaystyle{(k-1)\left(\frac{1}{2r}+ \frac{1}{2s} -\frac{k-1}{4rs}\right)\max_{u,v,i}{|h^{(i)}(u,v,p)|}}}.$$
\end{proof}

We finally present a bound on the condition number of the product ${\widehat{H}^TS\widehat{H}}$ as used in Section \ref{BNP_section}. Note that in general, $\kappa(AB) \le \kappa(A) \kappa(B)$ does not hold for rectangular matrices.

\noindent \textbf{Proposition 11} 
\textit{ Let $S$ be an $m\times m$ symmetric positive definite matrix and $\widehat{H}$ an $m\times n$ matrix. We have $\kappa({\widehat{H}^TS\widehat{H}}) \leq \kappa({\widehat{H}})^2\kappa({S})$.}

\begin{proof}
 We have $\displaystyle{\kappa(\widehat{H}^TS\widehat{H})=\frac{\lambda_{max}(\widehat{H}^TS\widehat{H})}{\lambda^*_{min}(\widehat{H}^TS\widehat{H})}}$ since $\widehat{H}^TS\widehat{H}$ is symmetric positive semi-definite, where $\lambda_{max}(\widehat{H}^TS\widehat{H})$ and $\lambda^*_{min}(\widehat{H}^TS\widehat{H})$ are the largest and the smallest nonzero eigenvalue of $\widehat{H}^TS\widehat{H}$ respectively. We only need to consider the case $ \lambda_{max}(\widehat{H}^TS\widehat{H})>0$. Using the Courant-Fisher minimax theorem  gives
\begin{align*}
    \lambda_{max}(\widehat{H}^TS\widehat{H}) &= \max_{x \perp \text{null}(\widehat{H})} \frac{x^T\widehat{H}^TS\widehat{H}x}{x^Tx} \\
    &= \max_{x \perp \text{null}(\widehat{H})} \frac{x^T\widehat{H}^TS\widehat{H}x}{x^T\widehat{H}^T\widehat{H}x} \frac{x^T\widehat{H}^T\widehat{H}x}{x^Tx} \\
& \leq \max_{x \perp \text{null}(\widehat{H})} \frac{(x^T\widehat{H}^T)S(\widehat{H}x)} {(x^T\widehat{H}^T)(\widehat{H}x)} \max_{x \neq 0} \frac{x^T\widehat{H}^T\widehat{H}x}{x^Tx}\\
& \leq \lambda_{max}(S) \lambda_{max}(\widehat{H}^T\widehat{H}),
\end{align*}
where we also impose $x\ne 0$ in all the maximizing sets above and the minimizing sets below. Similarly, we have
\begin{align*}
    \lambda^*_{min}(\widehat{H}^TS\widehat{H}) &= \min_{x \perp \text{null}(\widehat{H}^TS\widehat{H})} \frac{x^T\widehat{H}^TS\widehat{H}x}{x^Tx} \\
    &= \min_{x \perp \text{null}(\widehat{H}^TS\widehat{H})} \frac{x^T\widehat{H}^TS\widehat{H}x}{x^T\widehat{H}^T\widehat{H}x} \frac{x^T\widehat{H}^T\widehat{H}x}{x^Tx} \\
&\ge \min_{x \perp \text{null}(\widehat{H}^TS\widehat{H})} \frac{(x^T\widehat{H}^T)S(\widehat{H}x)}{(x^T\widehat{H}^T)(\widehat{H}x)} \min_{x \perp \text{null}(\widehat{H}^TS\widehat{H})} \frac{x^T\widehat{H}^T\widehat{H}x}{x^Tx}\\
&\ge \min_{\widehat{H}x \ne 0} \frac{(x^T\widehat{H}^T)S(\widehat{H}x)}{(x^T\widehat{H}^T)(\widehat{H}x)} \min_{x \perp \text{null}(\widehat{H})} \frac{x^T\widehat{H}^T\widehat{H}x}{x^Tx}\\
& = \lambda_{min}(S) \lambda^*_{min}(\widehat{H}^T\widehat{H}).
\end{align*}
Thus we have 

\begin{align*}
    \kappa(\widehat{H}^TS\widehat{H})&=\frac{\lambda_{max}(\widehat{H}^TS\widehat{H})}{\lambda^*_{min}(\widehat{H}^TS\widehat{H})}\\
    &\leq \frac{\lambda_{max}(S)\lambda_{max}(\widehat{H}^T\widehat{H})}{\lambda_{min}(S)\lambda^*_{min}(\widehat{H}^T\widehat{H})}\\
    &= \kappa(S) \kappa(\widehat{H})^2.
\end{align*}
\end{proof}

\textbf{\large Appendix B. Experimental Settings}

In this section, we list the detailed setting used in our experiments. Experiments were run using PyTorch 3 and Tensorflow versions 1.13.1 and 2.4.1. In particular, all fully connected networks and the 5-layer CNN use Tensorflow 1.13.1, with LN and BN with 4 Things run in Tensorflow  2.4.1. All ResNets and the exploratory experiments use PyTorch 3.

\textbf{Fully Connected Networks:}
For all fully connected networks, each model is trained using stochastic gradient descent, and the best learning rates for each network can be found in Table  \ref{learning_rate_table_fully_connected}. 
For all fully connected networks weights are initialized using Glorot Uniform \citep{Glorot10} with the biases initialized as zero.

\begin{table}[ht]
    \centering
\scalebox{0.85}{\begin{tabular}{ |c|c|c|c|c|c| } 
 \hline
 &Vanilla  & BN  & BNP & BN Renorm & LN \\ \hline
 mini-batch size 60 MNIST&$10^{-1}$ &$5 \times 10^{-1}$ &$5 \times 10^{-1}$ & N/A& $10^{-1}$\\ \hline
  mini-batch size 60 CIFAR10& $10^{-2}$ &$5 \times 10^{-1}$ &$10^{-1}$& N/A &  $5 \times 10^{-1}$ \\ \hline
  mini-batch size 6 CIFAR10& $5 \times 10^{-3}$& $5 \times 10^{-2}$ & $5 \times 10^{-2}$  & $5 \times 10^{-2}$ & $5 \times 10^{-2}$\\ \hline
  mini-batch size 1 CIFAR10& $5 \times 10^{-4}$ & N/A &  $10^{-1}$ & N/A & $10^{-2}$\\  
 \hline
\end{tabular}}
    \caption{Best learning rates for fully connected networks.}
    \label{learning_rate_table_fully_connected}
\end{table}



\textbf{CNNs:}
All experiments with the 5-layer CNN use the stochastic gradient descent optimizer, except for the vanilla network with mini-batch size 128 and the BN 4 things, where the Adam optimizer performs better and is used. Note that Adam was used in the implementation of the network in \cite{TensorCNN}, but  SGD is comparable to Adam in all other cases. 
All models use the weight initializer Glorot Uniform and the zero bias initializer unless otherwise mentioned. 
All default hyperparameters are used for BNP and all learning rates are found in Table  \ref{learning_rate_table_cnn}.

\begin{table}[ht]
    \centering
\scalebox{0.75}{\begin{tabular}{ |c|c|c|c|c|c|c|c| } 
 \hline
 &Vanilla& BN  & BNP & BNP+GN&GN&BN w/ 4 Things& BN Renorm  \\ \hline
 5-layer CNN mini-batch size 128 CIFAR10&$10^{-1}$ &$10^{-1}$ &$10^{-1}$& N/A & N/A& $5 \times 10^{-3}$ & N/A \\ \hline
  5-layer CNN mini-batch size 2 CIFAR10& $10^{-3}$ &$10^{-3}$ &$10^{-2}$& N/A &N/A &$10^{-4}$ & $10^{-2}$\\ \hline
  
 5-layer CNN mini-batch size 1 CIFAR10& $10^{-3}$ & N/A & $10^{-1}$  & N/A& N/A & $10^{-3}$ &N/A  \\ \hline
 ResNet-110 CIFAR10& N/A& $10^{-1}$&  $10^{-1}$& $10^{-1}$ & $10^{-1}$&  N/A &N/A\\  \hline
 ResNet-110 Preactivation CIFAR100& N/A &$10^{-1}$&  $10^{-2}$& $10^{-1}$ & $10^{-1}$ & N/A &N/A \\  \hline 
ResNet-18 ImageNet & N/A& $10^{-1}$ & N/A &$10^{-1}$ & $10^{-1}$& N/A & N/A \\
\hline
\end{tabular}}
    \caption{Best learning rates for CNN's. }
    \label{learning_rate_table_cnn}
\end{table}
%

%

\textbf{ResNets:} For the ResNet-110 experiments on CIFAR datasets, we follow the settings of \citep{He15,He16}.  For all models, we use the data augmentation as in \cite{He15} and  use the momentum optimizer with momentum set to 0.9. We implement all parameter settings suggested in \cite{He15} for BN, with the exception that Preactivation ResNet-110 for CIFAR-100 follows the learning rate decay suggested in \cite{DBLP:journals/corr/HanKK16}. 
These include weight regularization of $1E-4$ and  a learning rate warmup with initial learning rate $0.01$ increasing to $0.1$ after 400 iterations. 
For networks with GN, we follow \cite{wu2018group} and replace all BN layers with GN. We use group size 4. 
For BNP+GN with CIFAR10, we use weight regularization of $1.5E-4$, the He-Normal weight initialization  scaled by 0.1, and group size  4 in GN. For BNP+GN with CIFAR100, we use weight regularization of $2E-4$, the He-Normal weight initialization  scaled by 0.4,  and  group size 4. GN and BNP+GN use a linear warmup schedule,  with initial learning rate $0.01$ increasing to $0.1$ over 1 or 2 epochs, tuned for each network.


For the ResNet-18 experiment with ImageNet, we follow the settings of \cite{10.5555/2999134.2999257}. All images are cropped to $224 \times 224$ pixel size from each image or its horizontal flip \cite{10.5555/2999134.2999257}. All models use  momentum optimizer with 0.9, weight regularization $1E-4$, except BNP+GN uses $8.5E-4$, a mini-batch size of 256 and train on 1 GPU.  
All models use an initial learning rate of $0.1$ which is divided by 10 at 30, 60, and 90 epochs. 
Both GN and BNP+GN use groupsize 32.

The best learning rates for all models are listed in Table  \ref{learning_rate_table_cnn}.

\vskip 0.2in
 \small
\bibliography{paper}

\end{document}